\documentclass[lettersize,journal]{IEEEtran}
\usepackage{amsmath,amsfonts}
\usepackage{array}
\usepackage[caption=false,font=normalsize,labelfont=sf,textfont=sf]{subfig}
\usepackage{textcomp}
\usepackage{stfloats}
\usepackage{url}
\usepackage{verbatim}
\usepackage{graphicx}
\usepackage{cite}
\usepackage{xcolor}
\usepackage{cite}
\usepackage{times}
\usepackage{epsfig}
\usepackage{graphicx}
\usepackage{amsmath}
\usepackage{amssymb}
\usepackage{epstopdf}
\usepackage{url}
\usepackage{multirow}
\usepackage{booktabs}
\usepackage{tabularx}
\usepackage[noend]{algpseudocode}
\usepackage{bm}

\begin{document}

\title{\textbf{3DM-WeConvene: Learned Image Compression with 3D Multi-Level Wavelet-Domain Convolution and Entropy Model}}

% \author{%
% IEEE Publication Technology,~\IEEEmembership{Staff,~IEEE,}%
% \thanks{This paper was produced by the IEEE Publication Technology Group. 
% They are in Piscataway, NJ.}%
\author{Haisheng~Fu,
        Jie~Liang*,
        Feng~Liang,
        Zhenman~Fang,
        Guohe~Zhang,
        Jingning~Han
\thanks{Haisheng~Fu, Jie~Liang and zhenman~Fang are with the School of Engineering Science, Simon Fraser University, Canada (e-mails: hfa23@sfu.ca, jiel@sfu.ca; zhenman@sfu.ca) (*Corresponding authors: Jie Liang).}% <-this % stops a space
\thanks{Feng~Liang and Guohe~Zhang are with the School of Microelectronics, Xi'an Jiaotong University, Xi'an, China.  (e-mails: fengliang@xjtu.edu.cn; zhangguohe@xjtu.edu.cn).}% <-this % stops a space
\thanks{Jingning~Han is with the Google Inc. (e-mail: jingning@google.com).}
\thanks{This work was supported by the Natural Sciences and Engineering Research Council of Canada (RGPIN-2020-04525), Google Chrome University Research Program, NSERC Discovery Grant RGPIN-2019-04613, DGECR-2019-00120, Alliance Grant ALLRP-552042-2020; CFI John R. Evans Leaders Fund; MITACS Elevate Postdoc grant; National Natural Science Foundation of China (No. 61474093), Industrial Field Project - Key Industrial Innovation Chain (Group) of Shaanxi Province (2022ZDLGY06-02).}
}

% The paper headers
\markboth{Journal of \LaTeX\ Class Files,~Vol.~14, No.~8, August~2021}
{Fu \MakeLowercase{\textit{et al.}}: Learned Image Compression with Multi-Level Wavelet-Domain Convolution and Entropy Model}

%\IEEEpubid{0000--0000/00\$00.00~\copyright~2021 IEEE}
\maketitle

\begin{abstract}
Learned image compression (LIC) has recently made significant advancements and started to outperform the traditional image compression methods. However, most existing LIC approaches primarily operate in the spatial domain and lack an explicit mechanism for reducing frequency‐domain correlations. To bridge this gap, we propose a novel framework that effectively integrates the low-complexity 3D multi-level Discrete Wavelet Transform (DWT) into some convolutional layers and the entropy coding to reduce both spatial and channel correlations, thereby enhancing frequency selectivity and rate-distortion (R-D) performance.

In our proposed 3D multi-level wavelet-domain convolution (\textbf{3DM-WeConv}) layer, we first use 3D multi-level DWT such as the 5/3 and 9/7 wavelets in JPEG 2000 to convert the data into wavelet domain. After that, convolutions with different sizes are applied to different frequency subbands. The data are then converted back to spatial domain via inverse 3D DWT. The 3DM-WeConv layer can be used at different layers of existing convolutional neural network (CNN)-based LIC schemes.

Moreover, we propose a 3D wavelet-domain channel-wise autoregressive entropy model (\textbf{3DWeChARM}), which performs slice-based entropy coding in the 3D DWT domain, and low-frequency (LF) slices are coded first to serve as priors for high-frequency (HF) slices. We also employ a two-step training strategy that first balances LF and HF rates and then fine-tunes the system with separate weightings.

Extensive experiments demonstrate that the proposed framework consistently outperforms state-of-the-art CNN-based LIC methods in terms of R-D performance and computational complexity, with more gains for higher-resolution images. In particular, on the Kodak, Tecnick 100, and CLIC test sets, our method achieves BD-Rate reductions of $-12.24\%$, $-15.51\%$, and $-12.97\%$ respectively over H.266/VVC. We further show that the proposed 3DM-WeConv module can be used as a universal tool to improve the performance of other applications, such as video compression, image classification, image segmentation, and image denoising. The proposed 3DWeChARM entropy coding can also be used in transformer-based image/video compression schemes. The source code is available at \url{https://github.com/fengyurenpingsheng/WeConvene}.
\end{abstract}


\section{Introduction} 
\label{sec:intro}

\IEEEPARstart{T}{raditional} image compression approaches such as JPEG~\cite{JPEG}, JPEG2000~\cite{JPEG2000}, and BPG~\cite{BPG} primarily consist of three components: linear transform, quantization, and entropy coding, which were derived based on the Digital Signal Processing (DSP) and Information Theory. In particular, linear transforms such as the Discrete Cosine Transform (DCT) and Discrete Wavelet Transform (DWT) play an important role in removing the redundancy of the input, via linear filters with carefully designed frequency responses.

Recently, deep learning-based image compression (LIC) has achieved better performance than traditional approaches. In LIC, the linear transform is replaced by the more powerful deep learning-based neural networks to learn a compact representation of the input. The encoding networks generally follow the variational autoencoder (VAE) architecture ~\cite{Variational}. Early LIC methods predominantly utilized convolutional neural networks (CNNs) in the VAE framework ~\cite{Variational,Joint,GLLMM,Lee_2020,Lee_2021,He_2021_CVPR,He_2022_CVPR,jiang2023mlic}. Recently, transformer-based architectures have also been employed in the LIC frameworks~\cite{Liu_2023_CVPR, zhu2022transformerbased, Qian2022_ICLR}. However, transformer demands higher computational resources than CNN.

Despite significant advancements, a major limitation of the state-of-the-art (SOTA) LIC schemes is that they mainly operate in the spatial domain, and are not explicitly designed to remove the redundancy of the latent representations via frequency response considerations. Therefore, there is a big gap between the traditional DSP approach and the learning-based approach. It is desired to introduce frequency response to LIC to further improve its performance.

In LIC, deep learning is also used in the entropy coding to learn the distribution more accurately. The entropy coding in LIC often incorporated complex context models, such as serial auto-regressive context models~\cite{Joint, cheng2020, GLLMM}, to improve compression performance. However, the serial context models introduce substantial computational overhead.

In \cite{channel}, a channel-wise auto-regressive entropy model (ChARM) is proposed, which minimizes sequential processing, and achieves faster decoding by exploiting channel-wise dependencies, thereby enabling parallelism compared to spatially auto-regressive models. Motivated by \cite{channel}, a more advanced channel-wise entropy model is  developed in \cite{Liu_2023_CVPR}. This model integrates parameter-efficient swin-transformer-based attention (SWAtten) modules through channel squeezing, further improving rate-distortion (R-D) performance in both PSNR and MS-SSIM metrics. 

These serial context models and parallelized entropy coding in LIC are still performed in the spatial domain, which is different from the traditional approach, where entropy coding is almost always performed in the frequency domain after the linear transform and quantization. This is another gap between the traditional and the learning approaches.

\begin{figure*}[!thp]
	\centering
		\includegraphics[scale=0.43]{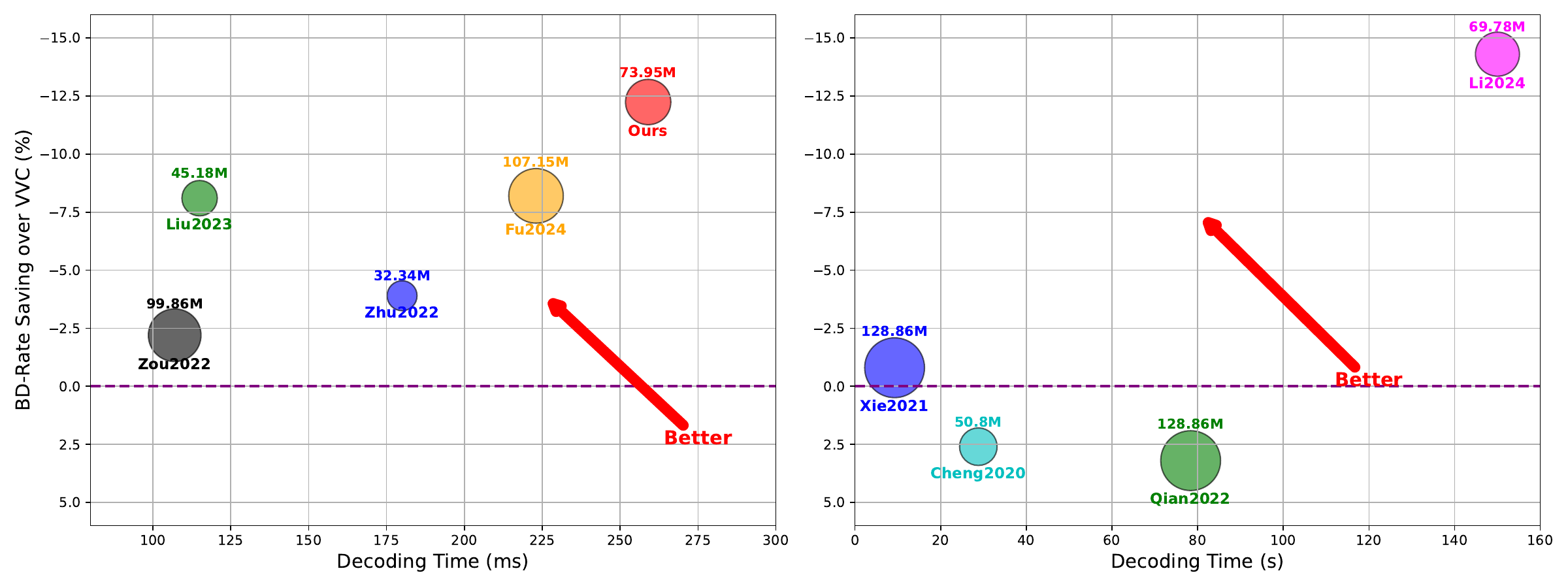}
	\caption{The decoding time, model size, and BD-Rate reductions over H.266/VVC for different LIC schemes on the Kodak test set. The area of each circle is proportional to the number of parameters (also written in the figure) of each model. Note that the unit of the left subfigure is milliseconds (ms), while the unit of the right subfigure is seconds (s). Our method achieves the best trade-off among the three metrics.}
	\label{BD_rate}
\end{figure*}

Several studies~\cite{Ma_2022_PAMI, Akyazi_2019_CVPR_Workshops, Lin_MMSP, Mahammand_AAAI} have tried to incorporate frequency-domain designs into LIC frameworks. However, these approaches have various limitations, and could not achieve the SOTA performance. 

In \cite{Fu_EECV}, an effective WeConvene approach is developed to integrate the classical DWTs such as the Haar, 5/3 and 9/7 wavelets into LIC, which achieves superior performance compared to traditional image codecs (e.g., H.266/VVC) as well as SOTA CNN-based LIC methods without increasing the complexity, since the complexity of the DWT is negligible compared to convolution layers. It proposes a Wavelet-domain Convolution (WeConv) layer and a Wavelet-domain Channel-wise Auto-Regressive entropy Model (WeChARM). WeConvene establishes a new paradigm by bridging the gaps between the traditional DSP and the new learning approaches for image compression, and enables more wavelet theories to be used in LIC.

In this work, we propose several improvements to WeConvene in \cite{Fu_EECV} and develop a 3D multi-level wavelet-domain convolution and entropy model (3DM-WeConvene) scheme.

The WeConv layer in \cite{Fu_EECV} only uses 2D spatial DWT. In this paper, we develop a 3D multi-level wavelet-domain convolution (3DM-WeConv) layer with three enhancements. First, we generalize the 2D DWT in WeConv to 3D DWT by also applying DWT across all channels, which effectively reduces the cross-channel correlations. Second, we extend the one-level DWT in WeConv to multi-level DWT, which can improve the frequency response and the R-D performance, especially for high-resolution images. Moreover, in WeConv, $3 \times 3$ convolutions are used in all wavelet subbands. Although this is a suitable choice for the low-frequency (LF) subband, it is not necessary for the high-frequency (HF) subbands, whose data are almost uncorrelated after DWT. In this paper, we only apply $3\times3$ convolution in the LF subband. For HF subbands, we apply the $1 \times 1$ convolutions, which help to preserve and further enhance the sparsity of these subbands, and also have smaller model size and lower complexity.
    
In the entropy coding part, we also extend the 2D DWT in \cite{Fu_EECV} to 3D DWT, which can improve the entropy coding efficiency. The DWT in the channel direction also allows us to use slice-based entropy coding across channels in the DWT domain, and encode LF slices first, which can then be used as priors to encode HF slices, thereby improving the R-D performance.

Since the HF subbands along the channel and spatial dimensions are sparser, less bits should be allocated to them. Therefore, we propose a two-step training method. First, equal weights are assigned to all frequency subbands to train the framework. After that, the system is fine-tuned by assigning more weights to the two LF subbands and less weights to other HF subbands in the loss function.

Compared to \cite{Fu_EECV} and other leading LIC methods, as shown in Fig. \ref{BD_rate}, the proposed 3DM-WeConvene method achieves better trade-off between R-D performance, decoding and complexity, without using high-complexity operators like transformers. On the Kodak, Tecnick 100, and CLIC test sets, our method achieves BD-Rate reductions of $-12.24\%$, $-15.51\%$, and $-12.97\%$ respectively over H.266/VVC. 
    
We also demonstrate that the proposed 3DM-WeConv layer can be used as a universal tool for other applications with improved performance, such as video compression, image classification, image segmentation, and image denoising. The proposed 3DWeChARM entropy coding can also be used in transformer-based image/video compression schemes. 

\section{Background and Related Work} 
\label{sec:related_work}

\subsection{Traditional Image Compression Techniques}

Traditional image compression methods, such as JPEG~\cite{JPEG}, JPEG2000~\cite{JPEG2000}, and video coding standards like H.264/AVC, H.265/HEVC, and H.266/VVC~\cite{VVC}, have been the cornerstone of digital image and video compression for decades. These methods typically use linear transforms, notably the Discrete Cosine Transform (DCT) and the Discrete Wavelet Transform (DWT), to decorrelate pixel data and concentrate energy into a few significant coefficients~\cite{JPEG2000}. Following the transform, quantization is applied to reduce the precision of these coefficients, effectively discarding less perceptually important information. Entropy coding is then used in the transform domain to remove remaining statistical redundancies, resulting in compact and efficient bitstreams. These key components were mainly derived from the DSP and Information Theory.

\subsection{Learned Image Compression}

Recently deep learning technologies have been introduced to the field of image compression. The learned Image Compression (LIC) approach typically employs nonlinear deep neural networks via the autoencoder framework to jointly optimize the encoder, decoder, and entropy model ~\cite{Variational, Joint}. By learning nonlinear transforms, LIC methods effectively capture complex patterns and dependencies in images, and have achieved superior compression efficiency compared to the traditional approach.

Early LIC models~\cite{end_to_end, GDN} primarily focus on replacing traditional linear transforms with convolutional neural networks (CNNs). The introduction of variational autoencoders (VAEs) with hyperpriors~\cite{Variational} facilitated the modeling of spatial dependencies within latent representations. Subsequent advancements incorporated more powerful entropy models, such as auto-regressive entropy models and Gaussian Mixture Model ~\cite{Joint, cheng2020, GLLMM, He_2022_CVPR}, as well as attention mechanisms ~\cite{Liu_2023_CVPR, cheng2020, GLLMM}.

However, these serial auto-regressive entropy models can only be decoded sequentially, resulting in high decoding complexity. To address this issue, parallelizable auto-regressive entropy models, such as the checkerboard entropy model~\cite{He_2021_CVPR, fu2023fast} and channel-wise auto-regressive entropy models (ChARM)~\cite{channel}, have been proposed to maximize decoding speed while maintaining compression performance.

Recently, transformer-based architectures have also been introduced into LIC frameworks~\cite{Liu_2023_CVPR, Qian2022_ICLR, FAT}, whicha can capture long-range dependencies within latent representations. Although transformer-based methods achieve superior compression performance, they often incur increased computational complexity and present significant challenges during training.

Maintaining a trade-off among compression performance, decoding efficiency, and model complexity is desired in image compression. In \cite{He_2022_CVPR}, uneven channel-conditional adaptive coding and an efficient model named ELIC are proposed to enhance coding performance and running speed. Similarly, in \cite{fu2023fast}, four techniques are introduced to balance the trade-off, achieving faster inference while retaining strong R-D performance.

\subsection{Frequency-Domain Processing in Learned Image Compression}

Motivated by the frequency-domain processing in the traditional DSP-based image processing, several methods have explored the integration of frequency-domain techniques into LIC frameworks. In \cite{Akyazi_2019_CVPR_Workshops}, a Daubechies-1 wavelet transform was first applied to the input image, and multiple CNN layers are then applied in the wavelet domain. In the decoder, after the decoder CNN, the inverse DWT is applied to recover the image. However, the performance of this wavelet-domain CNN approach was significantly lower than JPEG. Similarly, \cite{Iliopoulou2023} incorporated the 9/7 wavelet transform into a neural network architecture, but it also failed to achieve competitive results, highlighting the challenges of effectively integrating traditional wavelet transforms into LIC models.

Additionally, some methods have proposed wavelet-like transforms. For instance, several works~\cite{Mahammand_AAAI, Lin_MMSP, Mahammand_media, Fu_octave} have integrated the octave convolution~\cite{octave} into LIC, where the latent representations are divided into a low-resolution group and a high-resolution group, similar to the multi-resolution concept in DWT.

Gao et al.\cite{gao2021neural} proposed decomposing images into LF and HF components using pooling and subtraction operations, analogous to a Laplacian pyramid. This concept was extended to transformer-based LIC in \cite{Zafari_2023}, which divided multi-head self-attention modules into frequency-specific heads. In \cite{FAT}, a novel frequency-aware transformer (FAT) block is proposed, in which multiscale directional analysis is achieved via frequency-decomposition window attention and frequency-modulation feed-forward modules.

In ~\cite{Ma_2022_PAMI}, a lifting-like neural network architecture is proposed, which uses learned filters at each lifting step. 

All of these wavelet-like LIC methods still use learned filters, and do not employ well-established DWTs such as the 5/3 or 9/7 wavelets in JPEG 2000. The learned filters are not explicitly designed to have specific frequency responses.

The limitations of prior work highlight a gap in effectively leveraging frequency-domain processing within LIC frameworks. Our motivation is to bridge this gap by integrating existing DWT into both the autoencoder network and the entropy coding components of LIC, thereby removing frequency-domain correlations explicitly, improving the sparsity of the latent representations and the R-D performance, without introducing extra computational complexity. Our framework combines the benefits of the traditional DSP approach and the latest learning approach and will motivate more research in this direction.

\begin{figure*}[!thp] 
\centering 
\includegraphics[scale=0.5]{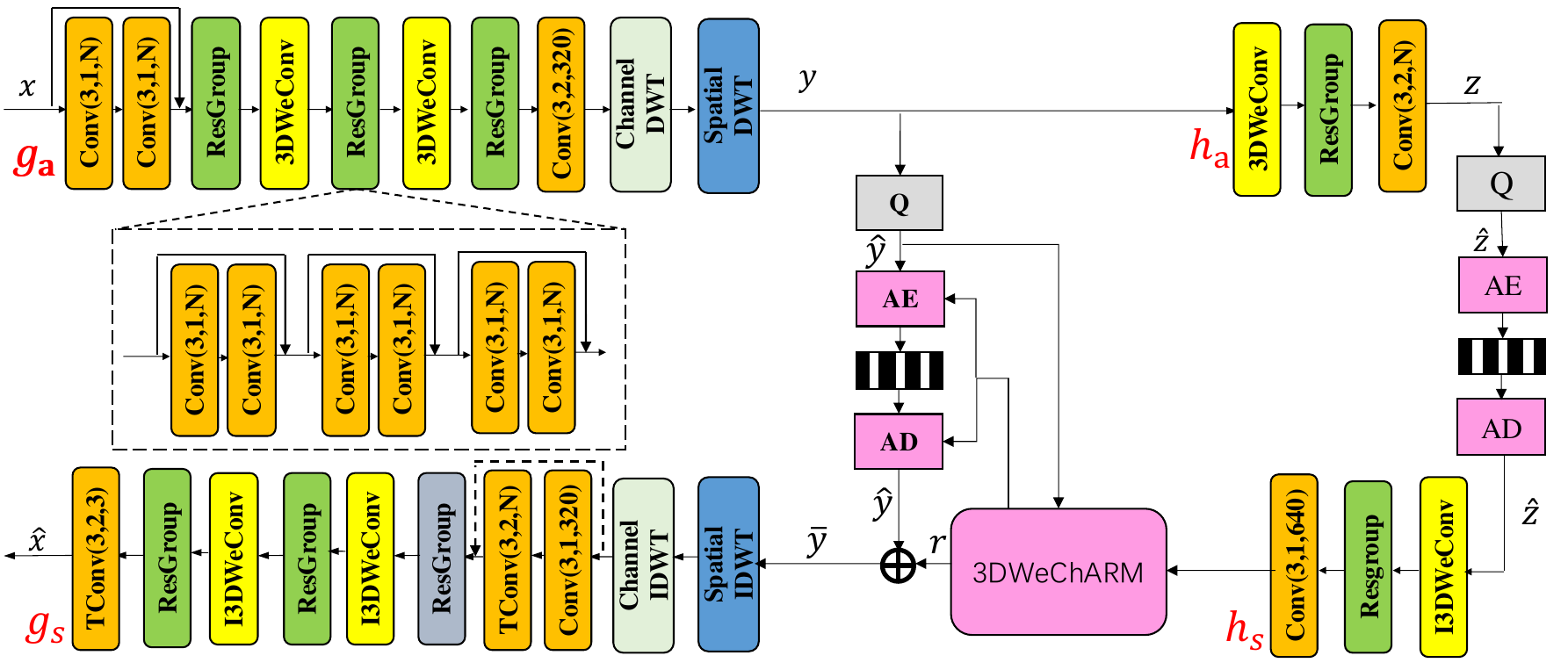} 
\caption{The overall architecture of the proposed 3DM-WeConvene scheme. The details of the 3DM-WeConv layer and the 3DWeChARM module are in Fig. \ref{fig:3DWeconv} and Fig. \ref{channel_wise_entropy_model} respectively. \texttt{Conv(3, s, N)} represents a convolutional layer with a $3 \times 3$ kernel size, stride $s$, and $N$ filters, while \texttt{TConv(3, s, N)} denotes a transposed convolutional layer. Dashed shortcut connections indicate changes in tensor size. The abbreviations \texttt{AE} and \texttt{AD} refer to the Arithmetic Encoder and Arithmetic Decoder in entropy coding, respectively.} 
\label{whole_networkstructure} 
\end{figure*}

\begin{figure}[!thp] 
\centering 
    \includegraphics[scale=0.37]{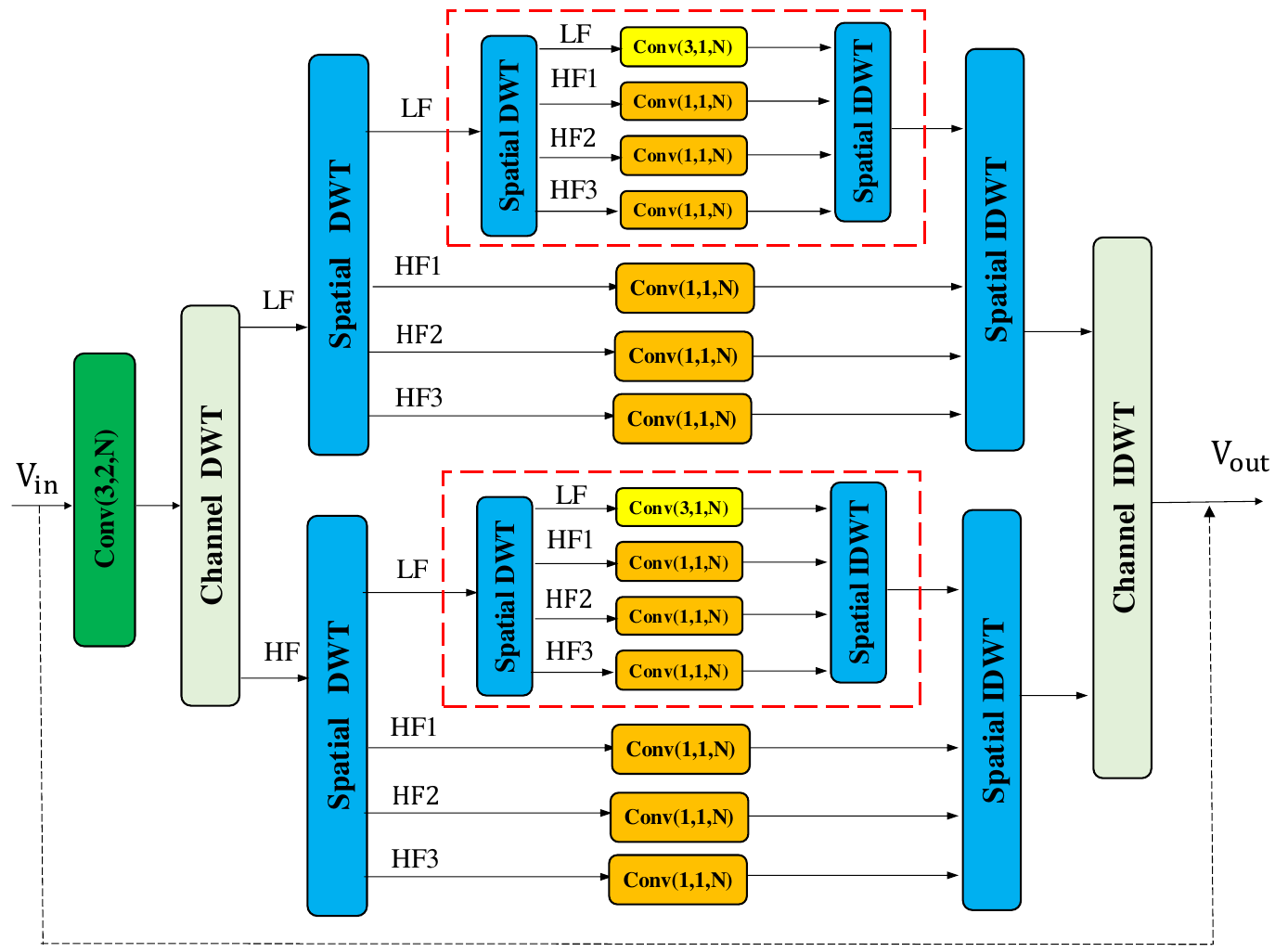} 
\caption{Forward 3DM-WeConv layer with downsampling.} 
\label{fig:3DWeconv} 
\end{figure}

\begin{figure}[!thp] 
\centering 
\includegraphics[scale=0.6]{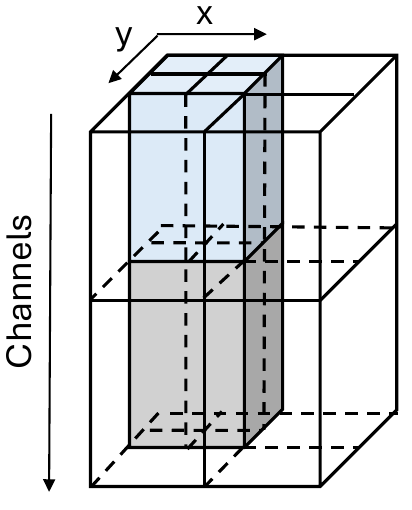} 
\caption{The 3D illustration of the 3DM-WeConv layer in Fig. \ref{fig:3DWeconv}.} 
\label{fig:3DWeconv-cube} 
\end{figure}

\section{The Architecture of the Proposed Scheme} 
\label{sec:proposed_method}

In this section, we present the overall architecture of our proposed method, which extends our previous work in \cite{Fu_EECV}. The two key components are the 3D multi-level wavelet-domain convolution (3DM-WeConv) layer and the 3D wavelet-domain channel-wise auto-regressive entropy model (3DWeChARM).

Fig.\ref{whole_networkstructure} illustrates the architecture of our framework. Similar to the prevailing LIC framework \cite{cheng2020,GLLMM,Liu_2023_CVPR,Fu_EECV}, our method includes a core encoder $g_a$ that extracts compact and efficient latent representations from the input image, a core decoder $g_s$ for image reconstruction, and hyperprior networks $h_a$ and $h_s$ that encode and decode side information to facilitate entropy coding of the latent representations.

The input $x$ is a color image of dimensions $W \times H \times 3$, with pixel values normalized to the range $[0, 1]$. The core encoder network $g_a$ comprises of multiple CNN layers. It includes some ResGroup layers, which further contains various ResNet blocks. The proposed 3DM-WeConv layer is employed at selected layers of the core networks and the hyperprior networks. The detailed explanation of the 3DM-WeConv layer are provided in Sec. \ref{sec_3DWeconv}.

In addition to the 3DM-WeConv layer, our framework also incorporates 3D DWT at the beginning of the entropy coding, via a 1D DWT across all channels and a spatial 2D DWT in each channel. The 3D DWT not only increases the sparsity of the latent representations, but also enables the proposed improved 3DWeChARM module in entropy coding, which encodes LF subbands first, and then uses them as priors to encode other HF subbands, as detailed in Sec. \ref{sec:3Dwecharm}.

\subsection{3D Wavelet-Domain Convolution Layer (3DM-WeConv)}
\label{sec_3DWeconv}

Fig.~\ref{fig:3DWeconv} shows an example of the proposed forward 3DM-WeConv layer. The input feature map $\mathbf{V}_{\text{in}} \in \mathbb{R}^{C \times H \times W}$ first undergoes a convolutional layer with an optional downsampling operator, where $C$, $H$, and $W$ are the number of channels, the height and the weight of the input. The corresponding inverse 3DM-WeConv (I3DM-WeConv) layer in the decoder network has the same structure, except that the first convolutional layer is replaced by the transposed convolution layer.

The proposed 3DM-WeConve layer has three enhancements over the WeConv layer in \cite{Fu_EECV}. It provides more fine-grained frequency-selective decomposition across spatial and channel directions, thereby improving the representational power and efficiency of learned features.

First, we generalize the 2D DWT in WeConv to 3D DWT by first applying a 1D DWT across all channels, which effectively reduces the cross-channel correlations. The classic DWT such as the Haar wavelet, and the 5/3 and 9/7 wavelets in JPEG 2000 can be used directly. In the example in Fig.~\ref{fig:3DWeconv}, we only use one-level 1D DWT in the channel direction, but more levels of channel DWT can be employed, depending on the applications.

After the channel DWT, all the channels are partitioned into two groups.

Next, 2D spatial DWT is applied to each channel in the two channel groups. In \cite{Fu_EECV}, only one-level of 2D DWT is used. In this paper, we generalize it to multi-level 2D DWT. In Fig.~\ref{fig:3DWeconv}, for simplicity purpose, only two-level 2D DWT is illustrated. In general, multi-level 2D DWT can be used for better performance, especially for high-resolution images, similar to JPEG 2000, as will be shown in the experimental result part.

In Fig.~\ref{fig:3DWeconv}, the first level of spatial 2D DWT creates four subbands. After that, another level of spatial 2D DWT is applied to the LF subband, leading to seven subbands in each channel.

The spatial-channel DWT subband partitions in Fig.~\ref{fig:3DWeconv} can also be visualized in Fig. \ref{fig:3DWeconv-cube}.

Another improvement of this paper over \cite{Fu_EECV} is that $3 \times 3$ convolutions are used in all wavelet subbands in \cite{Fu_EECV}. However, it is known that the coefficients in the HF subbands of DWT output are highly uncorrelated and thus very sparse. Therefore applying $3 \times 3$ convolutions to the HF subbands could destroy the sparsity of these subbands, if not designed properly. In this paper, we only apply $3\times3$ convolution in the LF subband. For HF subbands, we apply the $1 \times 1$ convolutions, which filter only across different channels, thus preserving and further enhancing the sparsity of these HF subbands. The $1 \times 1$ convolutions also have less complexity than the $3 \times 3$ convolutions. It will be shown in the ablation studies that $1 \times 1$ convolutions in HF subbands has better trade-off between performance and complexity than $3 \times 3$ convolutions.

This is also why the previously DWT-domain CNNs in \cite{Akyazi_2019_CVPR_Workshops,Iliopoulou2023} do not work well, because blindly applying multi-layer CNNs in the DWT domain could easily destroy the sparsity of the DWT output.
 
After these operations, inverse spatial and channel DWTs are applied to convert the signal back to the spatial domain. The 3DM-WeConv layer also has a standard residual connection to enhance the training stability.

Different from the previously DWT-based LIC efforts in \cite{Akyazi_2019_CVPR_Workshops,Iliopoulou2023}, the input and the output of our 3DM-WeConv layer are still in the spatial domain. The 3DM-WeConv layer only manipulates the data in the DWT domain. Therefore, the proposed 3DM-WeConv layer can be used as a standalone layer in CNN networks without drastically disrupting the typical signal distributions in spatial-domain CNNs.

By judiciously applying both spatial and channel DWTs, the proposed 3DM-WeConv layer obtains subbands with heightened sparsity, which in turn lowers the effective entropy of the representation and contributes to better R-D performance or more compact parameterization.

The 3DM-WeConv layer is beneficial not only for LIC but also for other tasks like video compression, classification, segmentation, and denoising, and wherever multi-scale feature extraction is desirable.

\subsection{3D Wavelet-Domain Channel-Wise Auto-Regressive Entropy Model (3DWeChARM)}
\label{sec:3Dwecharm}

\begin{figure}[!t]
\centering
\includegraphics[scale=0.4]{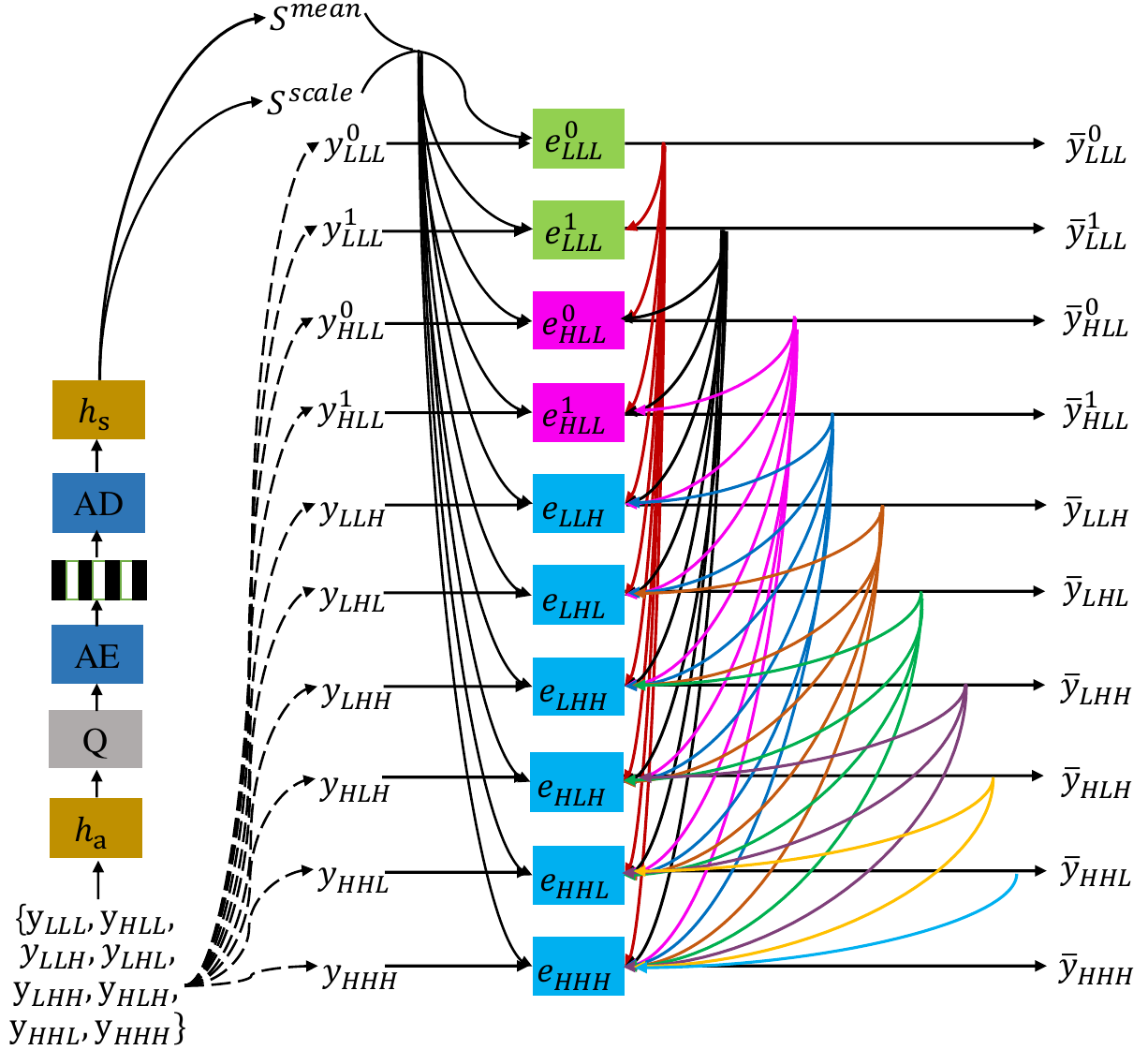}
\caption{Details of the 3DWeChARM module in entropy coding. The latent representation $y$ after 3D DWT is divided into slices, which are coded sequentially. The details of each slice coding are shown in Fig. \ref{fig:wecharm_structure}.}
\label{channel_wise_entropy_model}
\end{figure}

To further enhance compression efficiency, we generalize the 2D wavelet-domain channel-wise auto-regressive entropy model (WeChARM) in \cite{Fu_EECV} to 3D wavelet, as shown in Fig. \ref{channel_wise_entropy_model}

In \cite{Fu_EECV}, after 2D spatial DWT, the three HF subbands of all channels are concatenated into one HF subband. The channel slice idea in \cite{Liu_2023_CVPR} is then applied to divide all the channels in the LF subband into $5$ slices, each with 64 channels. All channels in the HF suband are also divided into $5$ slices, each with $192$ channels from three different HF subbands. The $10$ slices are then encoded sequentially, using the previously coded slices as prior.

In this paper, as shown in Fig. \ref{whole_networkstructure}, before entropy coding, we apply an one-level channel DWT and an one-level spatial DWT. This creates $8$ 3D DWT subbands, denoted as $y_{LLL}$, $y_{LLH}$, $y_{LHL}$, $y_{LHH}$, $y_{HLL}$, $y_{HLH}$, $y_{HHL}$, and $y_{HHH}$, where the first subscript represents LF or HF subband after the channel DWT, and the last two subscripts represent LF or HF subbands after the 2D spatial channel DWT.

To maintain a good balance between complexity and performance, we divide each of the two spatial LF subbands $y_{LLL}$ and $y_{HLL}$ into $2$ slices, each with $80$ channels, denoted by $y^0_{LLL}$, $y^1_{LLL}$, $y^0_{HLL}$, and $y^1_{HLL}$. Each of the other 6 HF subbands is treated as one slice, each with $160$ channels. As a result, the 3DWeChARM also has $10$ slices in total. 

Compared to the WeChARM in \cite{Fu_EECV}, although the number of slices is identical and the numbers of channels in the $10$ slices are also similar, the slices in 3DWeChARM are partitioned according to the frequency subbands; hence they are sparser than those in WeChARM, and the 3DWeChARM also does not group data from different HF subbands into one slice, thanks to the use of channel DWT. Therefore the estimation of the probability distribution parameters in each slice is more accurate than in WeChARM, because different DWT subbands have different distributions. This can improve the entropy coding efficiency.

These slices are then coded sequentially from LF to HF, using the previously coded slices as prior, as shown in Fig. \ref{channel_wise_entropy_model}, where a slice coding network $e_{Slice}$ is used to encode each slice. 

\begin{figure*}[!t]
\centering
\includegraphics[width=0.8\textwidth]{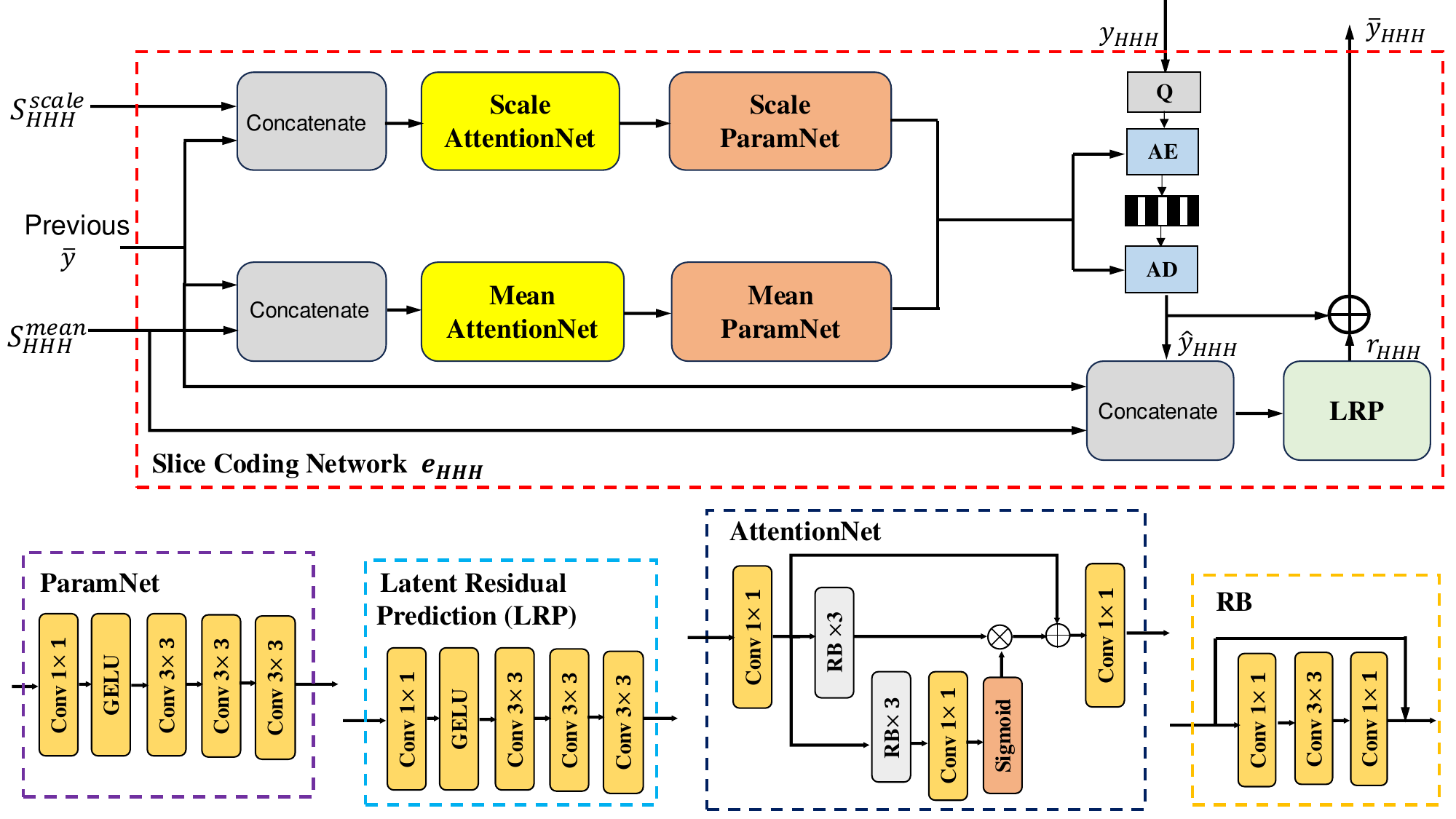}
\caption{An example of the slice coding network $e_{Slice}$ in Fig. \ref{channel_wise_entropy_model}.}
\label{fig:wecharm_structure}
\end{figure*}

The details of $e_{Slice}$ are shown in Fig. \ref{fig:wecharm_structure}, which is based on \cite{channel,Liu_2023_CVPR,Fu_EECV}. It uses the scale and mean parameters from the hyper-network, as well as previously coded slices to estimate the distribution parameters of each slice via the AttentionNet and ParamNet modules. It also uses the Latent Residual Prediction (LRP) module in \cite{channel} to predict the quantization error, and add it to the dequantized result to get the final reconstruction.

To reduce model parameters, we employ some $1 \times 1$ convolutions in the ParamNet and LRP modules.

\subsection{Loss Function and Training Procedure}

We train the proposed network end-to-end by first minimizing a R-D objective defined as:
\begin{equation}
    \mathcal{L} = \lambda \cdot D + R(\hat{\mathbf{y}}_{all}) + R(\hat{\mathbf{z}}),
\end{equation}
where $\hat{\mathbf{z}}$ represents the quantized hyperprior, and $R(\hat{\mathbf{y}}_{all})$ and $R(\hat{\mathbf{z}})$ denote the estimated bit rates for all DWT subbands and the hyperprior, respectively. The distortion $D$ can be measured using the mean squared error (MSE) between the original image $\mathbf{x}$ and the reconstructed image $\hat{\mathbf{x}}$, or the MS-SSIM metric.

After the training above, to better control bit allocation between DWT subbands, we introduce a rate reweighting mechanism in the loss function to fine-tune the networks:
\begin{equation}
    \begin{split}
        \mathcal{L} = & \ w_{\text{1}} \cdot (R_{\text{LLL}} + R_{\text{HLL}}) \\
        & + w_{\text{2}} \cdot (R_{\text{LLH}}+R_{\text{LHL}}+R_{\text{LHH}}+R_{\text{HLH}}+R_{\text{HHL}}+R_{\text{HHH}}) \\
        & + R(\hat{\mathbf{z}}) + \lambda \cdot D.
    \end{split}
\end{equation}
where $R_{\text{LLL}}$ and $R_{\text{HLL}}$ represent the estimated bit rates for the LLL and HLL subbands, respectively. $R_{\text{LLH}}$, $R_{\text{LHL}}$, $R_{\text{LHH}}$, $R_{\text{HLH}}$, $R_{\text{HHL}}$, $R_{\text{HHH}}$ represent the estimated bit rates for other subbands, respectively. 

By setting $w_{1} > w_{2}$, we explicitly encourage the network to allocate more bits to the two LF subbands, which usually have more energy.

The entire framework is trained via backpropagation. To facilitate gradient-based optimization, the non-differentiable quantization operation $Q(\cdot)$ is approximated using additive noise~\cite{Variational}.

Our models were optimized using both MSE and MS-SSIM metrics respectively. In the first training stage, for MSE optimization, we selected $\lambda$ values from the set $\{0.0025, 0.0035, 0.0067, 0.013, 0.025, 0.05\}$, each corresponding to a specific bit rate. For MS-SSIM optimization, $\lambda$ values were set to $\{5, 8, 16, 32, 64\}$. In both cases, the number of filters ($N$) for latent features was fixed at 128 across all rates.

In the first training stage, each model was trained for $1.5 \times 10^{6}$ iterations using the Adam optimizer, with a batch size of 8 and an initial learning rate of $1 \times 10^{-4}$. The learning rate was reduced by a factor of 10 every 100,000 iterations following the initial 750,000 iterations to ensure stable convergence. In the second training stage, each model is fine-tuned for 50 epochs with a fixed learning rate of e-4.

We prepared our training dataset by aggregating images from the CLIC~\cite{CLIC_test_2021}, LIU4K~\cite{LIU_dataset}, and COCO~\cite{coco_dataset} datasets. Images were resized to $2000 \times 2000$ pixels and augmented through rotation and scaling, resulting in 160,000 training image patches, each with a resolution of $480 \times 480$ pixels.

\begin{figure*}[!t]
\centering
\subfloat[]{\includegraphics[width=3.5in]{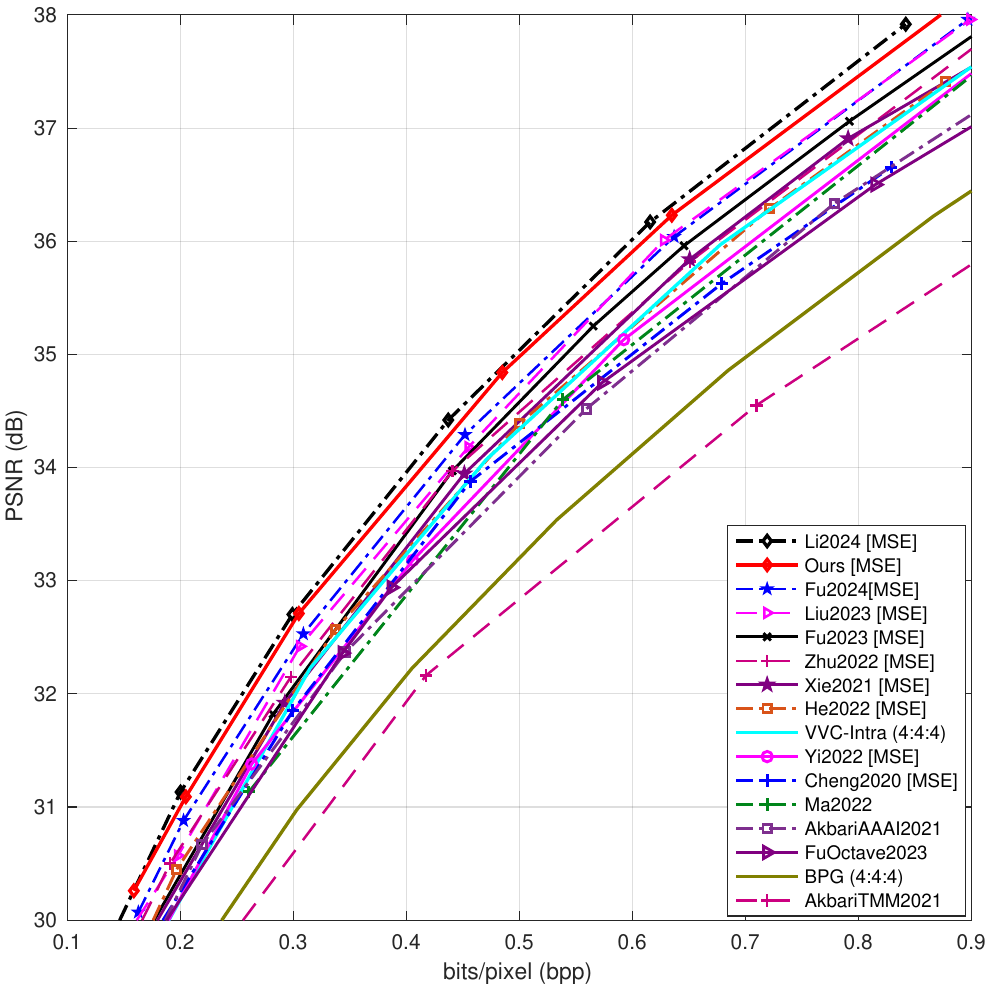}%
\label{fig_first_case}}
\hfil
\subfloat[]{\includegraphics[width=3.5in]{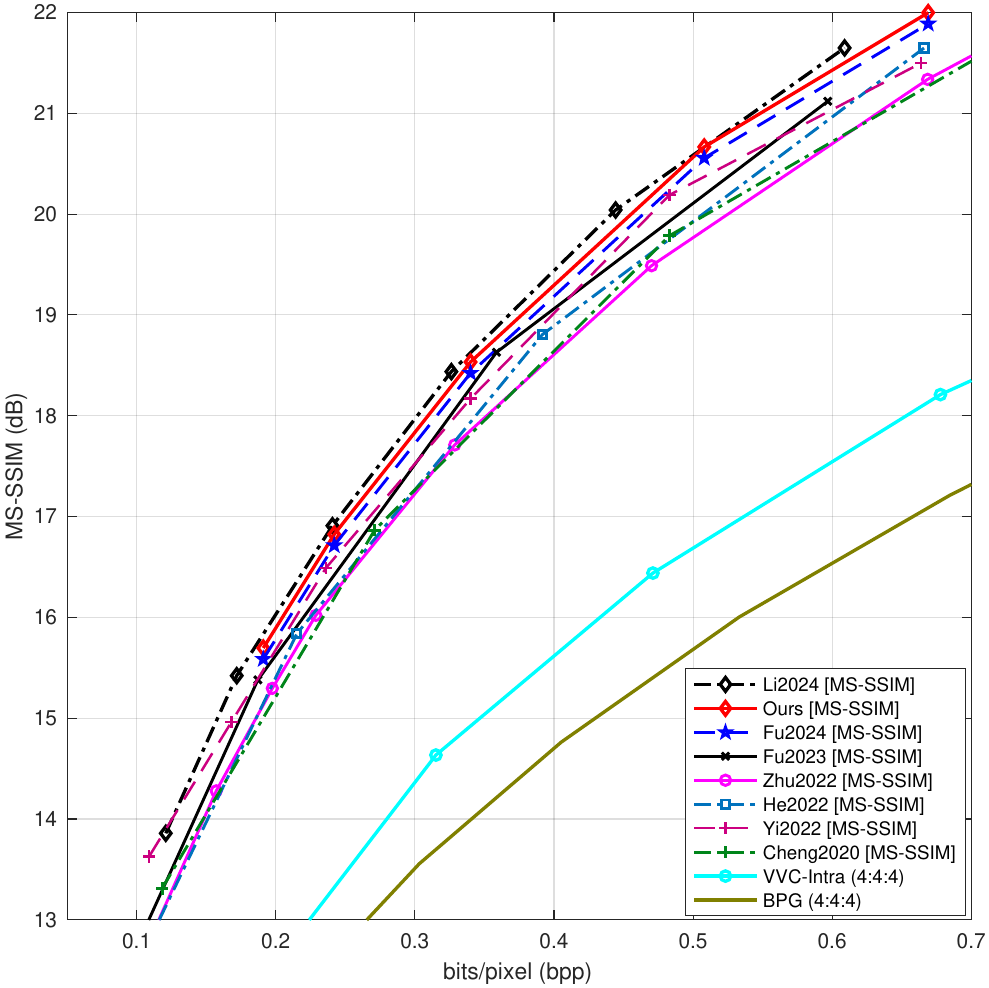}%
\label{fig_second_case}}
\caption{Average (a) PSNR and (b) MS-SSIM performances of different methods on the Kodak test set.}
\label{fig_Kodak}
\end{figure*}

\begin{figure*}[!t]
\centering
\subfloat[]{\includegraphics[width=3.5in]{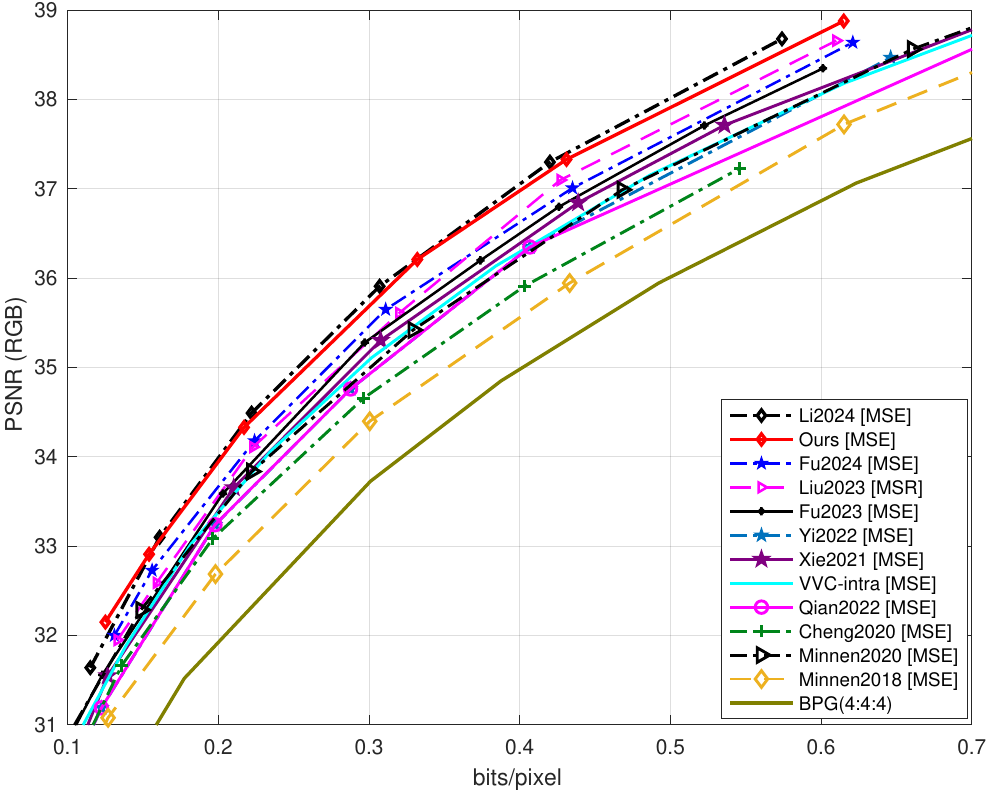}%
\label{fig_first_case}}
\hfil
\subfloat[]{\includegraphics[width=3.5in]{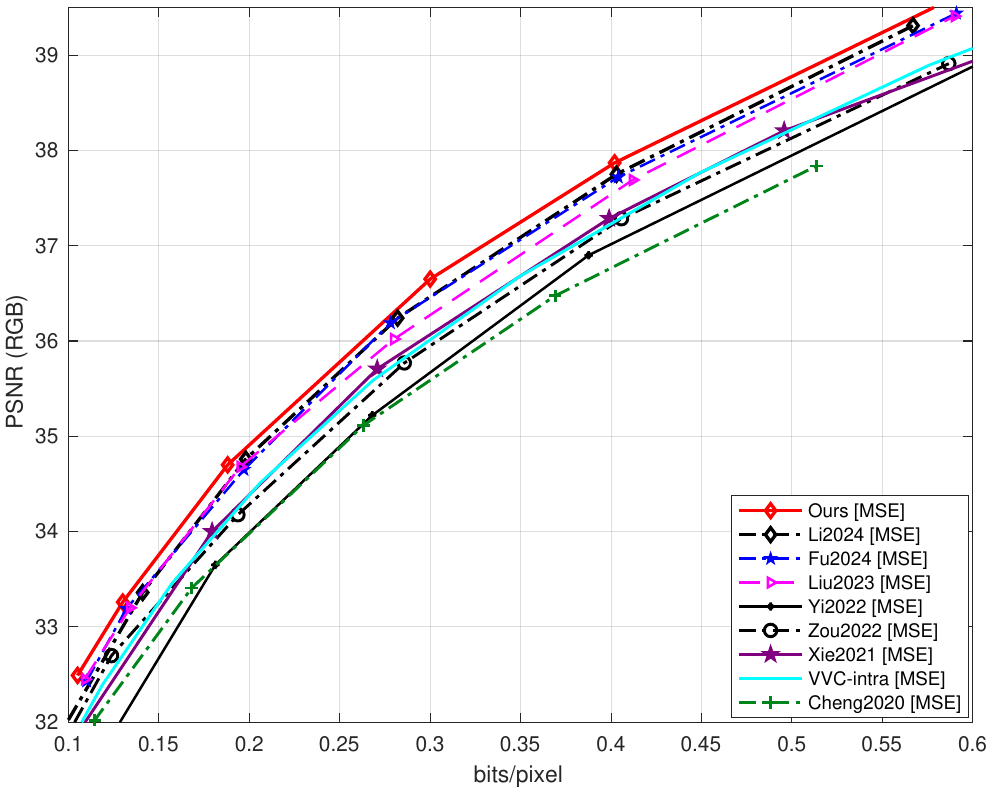}%
\label{fig_second_case}}
\caption{Average PSNR performances of different methods on (a) the Tecnick test set and (b) the CLIC test set.}
\label{fig_Tecnick_CLIC}
\end{figure*}

\subsection{Model Parameters}

For a standard 3D convolution with kernel size $K\times K$, $C_{\mathrm{in}}$ input channels, and $C_{\mathrm{out}}$ output channels, the number of parameters (including the biases) is given by \cite{Goodfellow-et-al-2016}:
\begin{equation}
   \label{eq:standard_conv_params_app}
   N_{\mathrm{conv}} = K^2 \cdot C_{\mathrm{in}} \cdot C_{\mathrm{out}} + C_{\mathrm{out}}.
\end{equation}

The number of model parameters in the proposed 3DM-WeConv layer and 3DWeChARM module can be obtained by modifying the formula above using the appropriate parameters in each subband.

\section{Experimental Results} \label{sec_results}

In this section, we evaluate the performance of our proposed method by comparing it with state-of-the-art LIC methods and traditional codecs, utilizing both Peak Signal-to-Noise Ratio (PSNR) and Multi-Scale Structural Similarity Index (MS-SSIM) as evaluation metrics. The LIC methods include Li2024 \cite{FAT}, Fu2024 \cite{Fu_EECV}, Liu2023 \cite{Liu_2023_CVPR}, Fu2023 \cite{GLLMM}, Zhu2022 \cite{zhu2022transformerbased}, Yi2022 \cite{Qian2022_ICLR}, He2022 \cite{He_2022_CVPR}, He2021 \cite{He_2021_CVPR}, Xie2021 \cite{xie2021enhanced}, AkbariAAAI2021 \cite{Mahammand_AAAI}, AkbariTMM2021 \cite{Mahammand_media}, Cheng2020 \cite{cheng2020}, Minnen2020 \cite{channel}, and Minnen2018 \cite{Joint}. Traditional codecs like H.266/VVC Intra (4:4:4) and H.265/BPG Intra (4:4:4) are also included for comparison.

We employ three widely recognized test sets: the Kodak PhotoCD test set \cite{Kodak} (24 images at resolutions of $768 \times 512$ or $512 \times 768$), the Tecnick 100 test set \cite{Tecnick} (100 images at $1200 \times 1200$ resolution), and the CLIC 2021 test set \cite{CLIC_test_2021} (60 images with resolutions ranging from $751 \times 500$ to $2048 \times 2048$), which has more high-resolution images.

The results for the learned methods are obtained from their pretrained models or reported results in their respective papers. Since some models are not evaluated on certain test sets or are not trained with the MS-SSIM metric, some methods are excluded from our comparisons. For Liu2023 \cite{Liu_2023_CVPR}, as only small pretrained models are available, we test these models on different test sets and report their model parameters, as well as encoding and decoding times. For Fu2024 \cite{Fu_EECV} and our proposed method, we present the results using the 9/7 wavelet.

\subsection{R-D Performance}

Figure~\ref{fig_Kodak} illustrates the average R-D curves for various methods on the Kodak test set, evaluated using PSNR and MS-SSIM metrics. Among the PSNR-optimized approaches, Li2024 \cite{FAT} achieves the best R-D performance among the learned methods and outperforms H.266/VVC across all bit rates, but its complexity is quite high, because it uses the more complicated transformer architecture. Our proposed method ranks the second, and is only up to 0.1 dB lower than Li2024 \cite{FAT}, with smaller gap at low rates.  We will show in Sec. \ref{sec_speed} that the complexity of Li2024 \cite{FAT} is more than 600 times higher than our method. Compared to our previous work in \cite{Fu_EECV}, the proposed method has a marked improvement of 0.2-0.3 dB, thanks to the 3DM-WeConv layer and the 3D DWT-based entropy coding.

For the MS-SSIM metric, Li2024 \cite{FAT} also achieves the best performance. Our proposed method has smaller gap with Li2024 than the PSNR metric, and outperforms \cite{Fu_EECV}, other LIC methods, and VVC at all rates.

Figure \ref{fig_Tecnick_CLIC}(a) presents the PSNR performance on the Tecnick 100 test set. Our method achieves almost identical performance to Li2024 \cite{FAT}, and has up to 0.5 dB gain over \cite{Fu_EECV}.

In Figure \ref{fig_Tecnick_CLIC}(b), we compare the PSNR performance on the CLIC 2021 test set, which has higher-resolution images (the longer side of each image is 2048 pixels). In this case, our proposed method achieves better performance than Li2024 \cite{FAT} at all rates (up to 0.2 dB), indicating that 3D and multi-level DWT is very suitable for high-resolution images. Even our previous work in Fu2024 \cite{Fu_EECV} has almost identical performance to Li2024 \cite{FAT}, and is better than other methods. It is expected that the performance of the proposed method for higher-resolution images such as 4K or 8K images can be further improved by using more levels of 3D DWT. 

\subsection{Trade-Off between Performance and Speed}
\label{sec_speed}

\begin{table}[!t]
\caption{Comparisons of encoding/decoding time, BD-Rate reduction over VVC, and model sizes of the low bit rates and high bit rates for the Kodak test set.}
\begin{center}
\begin{tabular}{ccccc}
\hline
\textbf{Methods} & \textbf{Enc.} & \textbf{Dec.} & \textbf{BD-Rate} &\textbf{$\#$Params}\\ 
\hline

\multicolumn{5}{c}{\textbf{Kodak}}\\
\hline
VVC  & 402.3s & 0.61s & 0.0 & -\\
Cheng2020 \cite{cheng2020}  &27.6s &28.8s &2.6\% & 50.80 MB \\
Hu2021 \cite{Hu_2021}  &32.7s &77.8s &11.1\% & 84.60 MB \\
He2021 \cite{He_2021_CVPR}  &20.4s &5.2s &8.9\% & 46.60 MB \\ 
Xie2021 \cite{xie2021enhanced} &4.10s &9.25s &-0.8\% & 128.86 MB\\
Zhu2022 \cite{zhu2022transformerbased} &0.27s &0.18s &-3.9\% & 32.34 MB\\
Zou2022 \cite{Zou_2022_CVPR} &0.106s &0.107s &-2.2\% & 99.86 MB\\
Qian2022 \cite{Qian2022_ICLR} &1.5 s &78.37s &3.2\% & 128.86 MB\\
Fu2023 \cite{GLLMM}  &420.6s &423.8s & -3.1\% & - \\ 
Liu2023\cite{Liu_2023_CVPR} &0.100 s &0.104s & -8.1\% &76.57 MB \\
Fu2024 \cite{Fu_EECV}  &0.222s & 0.223s & -9.8\% &  113.46 MB\\
%\textbf{Ours (Haar)} & \textbf{0.216s} & \textbf{0.218s} & \textbf{-9.84\%} & \textbf{63.01 MB}\\ 
Li2024 \cite{FAT} & $>$150s & $>$150s & -14.38\% & 69.78 MB\\ 
\textbf{Ours} & \textbf{0.256s} & \textbf{0.259s} & \textbf{-12.24\%} & \textbf{73.95 MB}\\ 
\hline

\multicolumn{5}{c}{\textbf{Tecnick}}\\
\hline
VVC  & 700.59s & 1.49s & 0.0 & - \\
Zou2022 \cite{Zou_2022_CVPR} & 0.431s & 0.472s & -2.6\% & 99.9 MB\\
Liu2023\cite{Liu_2023_CVPR} & 0.286s & 0.280s & -8.34\% & 45.18 MB\\
Fu2024 \cite{Fu_EECV} & 0.766s & 0.672s & -12.31\% & 113.46 MB\\
Li2024 \cite{FAT} & $>$630s & $>$630s & -16.15\% & 69.78 MB \\
\textbf{Ours}  & \textbf{0.823s} & \textbf{0.702s} & \textbf{-15.51\%} & \textbf{73.95 MB}\\
\hline
\multicolumn{5}{c}{\textbf{CLIC}}\\
\hline
VVC  & 949.58s & 1.98s & 0.0 & - \\
Zou2022 \cite{Zou_2022_CVPR} & 0.461s & 0.433s & 0.7855\% & 99.9 M\\
Liu2023\cite{Liu_2023_CVPR} &  0.480s & 0.445s & -7.68\% & 45.18 M\\
Fu2024 \cite{Fu_EECV} & 1.307s & 1.088s & -9.43\% & 113.46 MB\\
Li2024 \cite{FAT} & $>$10000s & $>$10000s & -9.81\% & 69.78 MB \\
\textbf{ours}  & \textbf{1.31s} & \textbf{1.14s} & \textbf{-12.97\%} & \textbf{73.95 M}\\ 

\hline
\end{tabular}
\label{BD_runing_time}
\end{center}
\end{table}

Table \ref{BD_runing_time} presents a comprehensive comparison of the average encoding and decoding times, BD-Rate reductions over H.266/VVC \cite{VVC}, and model sizes (estimated using the PyTorch Flops Profiler tool) for various LIC schemes on the Kodak, Tecnick, and CLIC test sets.  Some results for the Kodak test set are also shown in Fig. \ref{BD_rate}.

The results of our proposed method and Fu2024 \cite{Fu_EECV} are obtained using the 9/7 wavelet transform. The results of Liu2023 \cite{Liu_2023_CVPR} are obtained using their released small-sized models.

All experiments were executed on an NVIDIA Tesla 4090 GPU with 24~GB of memory, except for H.266/VVC, which was evaluated on a 2.9 GHz Intel Xeon Gold 6226R CPU. Note that the parameter count for Fu2023 \cite{GLLMM} is not reported due to its TensorFlow implementation, although \cite{GLLMM} indicates that its model complexity is considerably higher than that of Cheng2020 \cite{cheng2020}.

Several LIC methods, such as Cheng2020 \cite{cheng2020}, Fu2023 \cite{GLLMM}, Xie2021 \cite{xie2021enhanced}, Qian2022 \cite{Qian2022_ICLR}, and Li2024 \cite{FAT}, use serial auto-regressive entropy models, resulting in long encoding and decoding times. In contrast, recent LIC methods like Zhu2022 \cite{zhu2022transformerbased} and Zou2022 \cite{Zou_2022_CVPR} utilize GPU-friendly parallel auto-regressive entropy models to accelerate decoding while maintaining competitive R-D performance.

Our previous method \cite{Fu_EECV} achieves a BD-Rate reduction of $-9.8\%$, $-12.31\%$, and $-9.43\%$ relative to H.266/VVC on the three test sets respectively. Compared to this, the proposed approach in this paper further improves the BD-Rate reduction to $-12.24\%$, $-15.51\%$, and $-12.97\%$, with an average improvement of $3.06\%$. The running time increases about $10\%$ for the small-resolution Kodak set, and is almost identical for the high-resolution CLIC set. Moreover, the model size of the proposed method is actually $35\%$ less than \cite{Fu_EECV}.

Although Li2024 \cite{FAT} has better performance for the Kodak and Tecnick test sets, its complexity is more than 600 higher than our method. For the high-resolution CLIC test set, our method is $3.16\%$ better than Li2024 \cite{FAT}, and also more than 10,000 times faster than.

In other LIC methods, although Liu2023 \cite{Liu_2023_CVPR} is faster and has less model parameters, its R-D performance is inferior to our method. Overall, our proposed scheme strikes a more favorable balance between decoding speed, model complexity, and compression performance, especially for higher-resolution images.

\subsection{Comparisons of Different DWT-based Convolution Layers} 
\label{Ablation}

\begin{figure}[!t]
\centering
\includegraphics[scale=0.4]{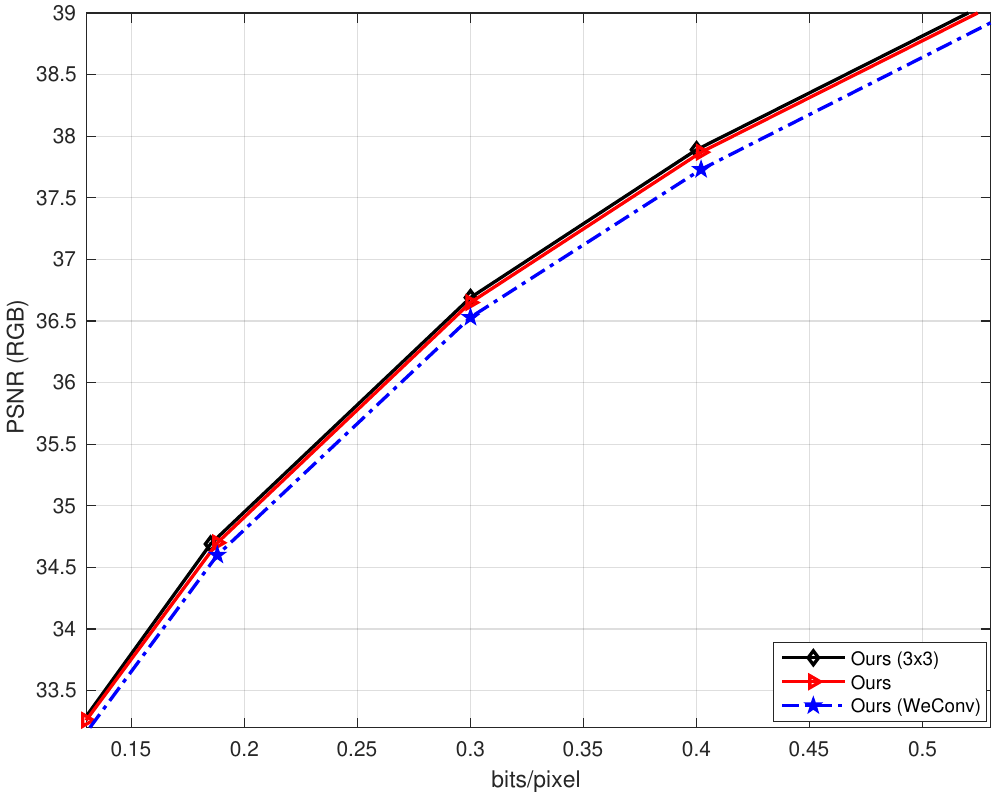}
\caption{R-D performance of different DWT-based convolution layers on the Kodak test set.}
\label{Fig_different_weconv_module}
\end{figure}

Fig. \ref{Fig_different_weconv_module}  and Table \ref{Table_comparsion_weconv} compare different DWT-based convolution layers on the Kodak test set, where “Ours” represents the proposed method with 3DM-WeConv module. "Ours (WeConv)" replaces the 3DM-WeConv module by the WeConv module from \cite{Fu_EECV}. "Ours ($3 \times 3$)" replaces the $1 \times 1$ convolutions in 3DM-WeConv by $3 \times 3$ convolutions (same as \cite{Fu_EECV}).

Fig. \ref{Fig_different_weconv_module} shows that the R-D performance of Ours (WeConv) is lower than Ours by 0.1-0.15 dB across all bit rates, demonstrating the advantage of the proposed 3DM-WeConv over WeConv. Ours ($3 \times 3$) has slightly better R-D performance than Ours. However, as shown in Table \ref{Table_comparsion_weconv}, the number of parameters of Ours is $13\%$ less than Ours ($3 \times 3$). Therefore using $1\times1$ convolutions in the HF subbands of DWT has better trade-off between complexity and performance than $3\times3$ convolutions.

\begin{figure*}[!t]
\centering
\subfloat[]{\includegraphics[width=3.0in]{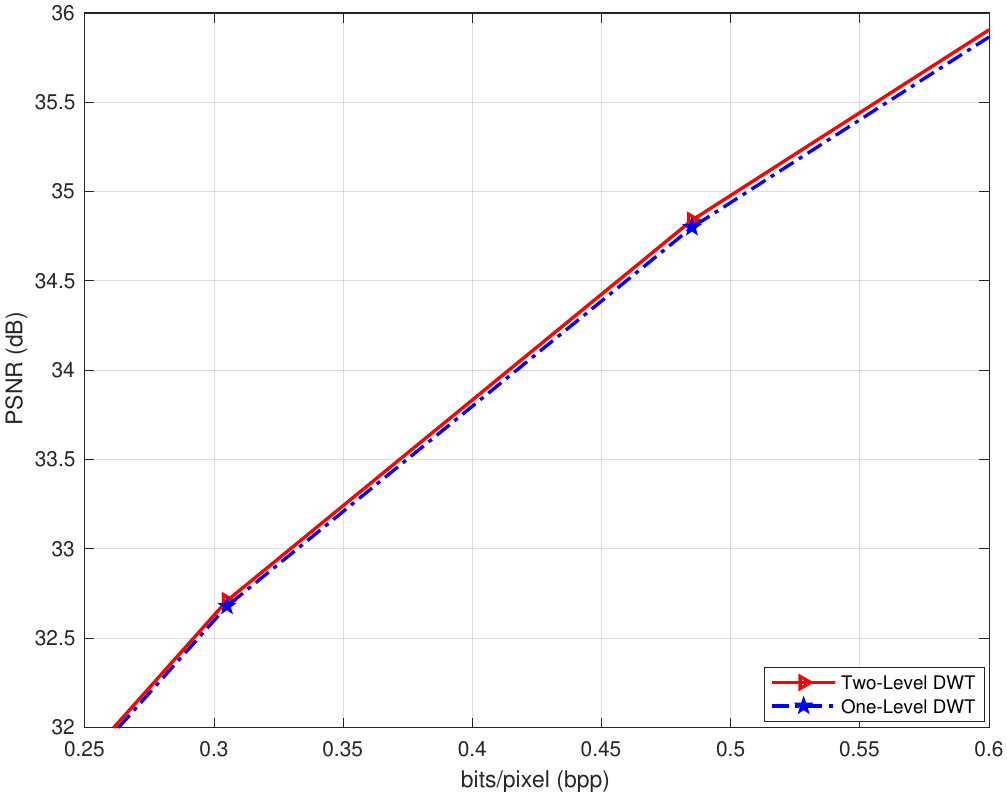}%
\label{fig_first_case}}
\hfil
\subfloat[]{\includegraphics[width=3.0in]{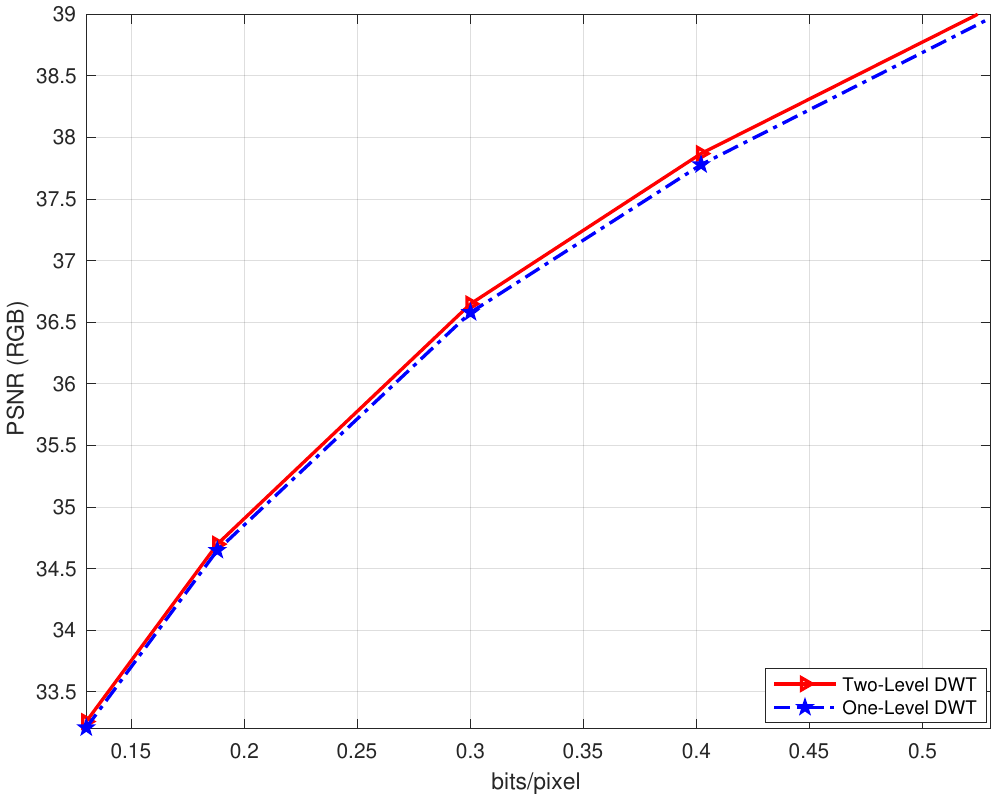}%
\label{fig_second_case}}
\caption{R-D performances of 3DM-WeConv layer with one and two levels of DWT. (a) Kodak test set. (b) CLIC test set.}
\label{fig_different_weconv}
\end{figure*}

\begin{table}[!t]
\caption{Comparisons of different DWT-based Convolution Layers in encoding/decoding time, and model sizes on Kodak test set.}
\begin{center}
\begin{tabular}{ccccc}
\hline
\textbf{Methods} & \textbf{Encoding time} & \textbf{Decoding time} &\textbf{$\#$Params}\\ 
\hline
\textbf{Ours} & \textbf{1.31s} & \textbf{1.14s} & \textbf{73.95 MB}\\ 
\textbf{Ours (WeConv)}  &\textbf{1.298s} & \textbf{1.111s} & \textbf{83.90 MB}\\
\textbf{Ours ($3 \times 3$)}  &\textbf{1.32s} & \textbf{1.18s} & \textbf{82.89 MB}\\
\hline

\end{tabular}
\label{Table_comparsion_weconv}
\end{center}
\end{table}

\subsection{Impact of Levels of DWT in 3DM-WeConv} 
\label{sec:wavelet_levels}

In this subsection, we compare the performance of 3DM-WeConv layer with one or two levels of spatial DWT, without changing any other component of the system. The results of the Kodak and CLIC test sets are presented in Figure \ref{fig_different_weconv} and Table~\ref{Table_different_Levels}.

It can be seen that using two-level spatial DWT can improve the R-D performance by 0.02–0.03 dB on the Kodak test set and 0.03–0.05 dB on the CLIC test set. This shows that multi-level DWT can enhance image compression performance, with more gains for higher-resolution images. Moreover, two-level DWT only increases the encoding time by $3\%$, decoding time by $7\%$, and model parameters by $1\%$.

It is expected that for higher-resolution images (e.g., 4K or 8K), applying more levels of DWT can further enhance the compression performance.

\begin{table}[!t]
\caption{Encoding time, decoding time, and model size of 3DM-WeConv layer with one and two levels of DWT for Kodak test set.}
\begin{center}
\begin{tabular}{cccccc}
\hline
\textbf{Test Set}&\textbf{DWT Levels} & \textbf{Encoding} & \textbf{Decoding} &\textbf{$\#$Params}\\ 
\hline
\textbf{Kodak}&\textbf{One} &\textbf{0.247s} & \textbf{0.240s} &  \textbf{72.25 MB}\\
\textbf{}& \textbf{Two} & \textbf{0.256s} & \textbf{0.259s} & \textbf{72.95 MB} \\
\hline
\textbf{CLIC}&\textbf{One} &\textbf{1.280s} & \textbf{1.084s} &  \textbf{72.25 MB}\\
\textbf{}& \textbf{Two} & \textbf{1.31s} & \textbf{1.14s} & \textbf{72.95 MB} \\
\hline
\end{tabular}
\label{Table_different_Levels}
\end{center}
\end{table}

\subsection{Comparison between 3DWeChARM and WeChARM} 
\label{Ablation}

\begin{figure}[!t]
\centering
\includegraphics[scale=0.5]{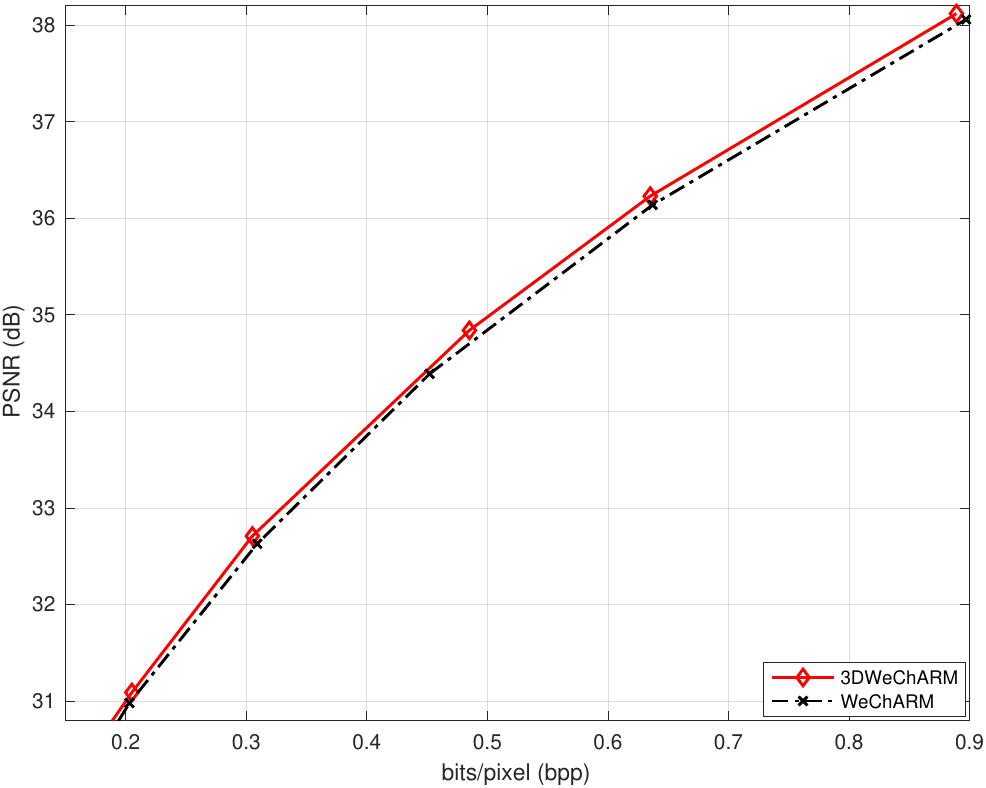}
\caption{R-D performance comparison between 3DWeChARM and WeChARMs on the Kodak test set.}
\label{abalation_wecharm}
\end{figure}

\begin{table}[!t]
\caption{Comparison of encoding time, decoding time, and model sizes when using 3DWeChARM and WeChARM on the Kodak test set.}
\begin{center}
\begin{tabular}{cccc}
\hline
\textbf{Methods} & \textbf{Encoding time} & \textbf{Decoding time}  &\textbf{$\#$Params}\\ 
\hline
\textbf{WeChARM}  &\textbf{0.210s} & \textbf{0.222s} &  \textbf{97.19 MB}\\
\textbf{3DWeChARM} & \textbf{0.256s} & \textbf{0.259s} & \textbf{72.95 MB}\\ 
\hline
\end{tabular}
\label{abalation_wecharm_table}
\end{center}
\end{table}

In this part, we report the improvement of 3DWeChARM over WeChARM in the entropy coding part. To ensure a fair comparison, we only replace the 3DWeChARM module in our framework by the WeChARM in \cite{Fu_EECV} while keeping other components unchanged. 

Fig. \ref{abalation_wecharm} shows the results using the Kodak test set, which show that 3DWeChARM has 0.1-0.15 dB gain over WeChARM at all bit rates. 

Table \ref{abalation_wecharm_table} compares the encoding time, decoding time, and model complexity, which shows that 3DWeChARM increases the encoding time by $22\%$, decoding time by $17\%$, but the model parameters are reduced by $25\%$.

\begin{table}[ht]
\centering
\caption{Bit‐rate (bpp) and percentage of all bits for different subbands and hyperprior signal $\hat{z}$, PSNR, and MS-SSIM when 3DWeChARM and WeChARM are used for the \texttt{kodak\_12} image. }
\begin{tabular}{lcc}
\hline
\textbf{Components} & 
\textbf{3DWeChARM} & 
\textbf{WeChARM}  \\
\hline
LLL          & 0.0802 / 24.1\% & —  \\
HLL          & 0.0597 / 17.9\% & — \\
LLH          & 0.0539 / 16.2\% & — \\
LHL          & 0.0473 / 14.2\% & — \\
LHH          & 0.0278 /  8.3\% & — \\
HLH          & 0.0169 /  5.1\% & — \\
HHL          & 0.0162 /  4.9\% & — \\
HHH          & 0.0130 /  3.9\% & — \\
\hline
LL\      & —               & 0.1054 / 31.4\% \\
LH           & —               & 0.0791 / 23.6\% \\
HL           & —               & 0.0720 / 21.5\% \\
HH           & —               & 0.0607 / 17.9\% \\
%y            & —               & —               \\
\hline
$\hat{z}$            & 0.0179 /  5.4\% & 0.0175 /  5.2\%  \\
\hline
\textbf{Total Bit-rate (bpp)} 
& 0.333          & 0.335                   \\
\textbf{PSNR (dB)}          
& 37.23          & 37.17                   \\
\textbf{MS-SSIM (dB)}       
& 17.08          & 17.08                   \\
\hline
\end{tabular}
\label{tab:keda12_comparison}
\end{table}

Table \ref{tab:keda12_comparison} shows the bit-rates of different subbands and the hyperprior signal $\hat{z}$ when 3DWeChARM and WeChARM are used for the \texttt{kodak\_12} image, which shows that after the channel DWT in 3DWeChARM, the LLL and HLL subbands have the highest bit rates. Therefore they are coded first in the entropy coding. The LHH, HLH, HHL, and HHH subbands are very sparse, which are easier to compress. Without channel DWT, WeChARM only has 4 subbands, whose bit rates are more similar to each other. This explains the improved R-D performance of 3DWeChARM over WeChARM.

\subsection{The effect of two-stage training method} 
\label{Ablation}

\begin{figure}[!t]
\centering
\includegraphics[scale=0.4]{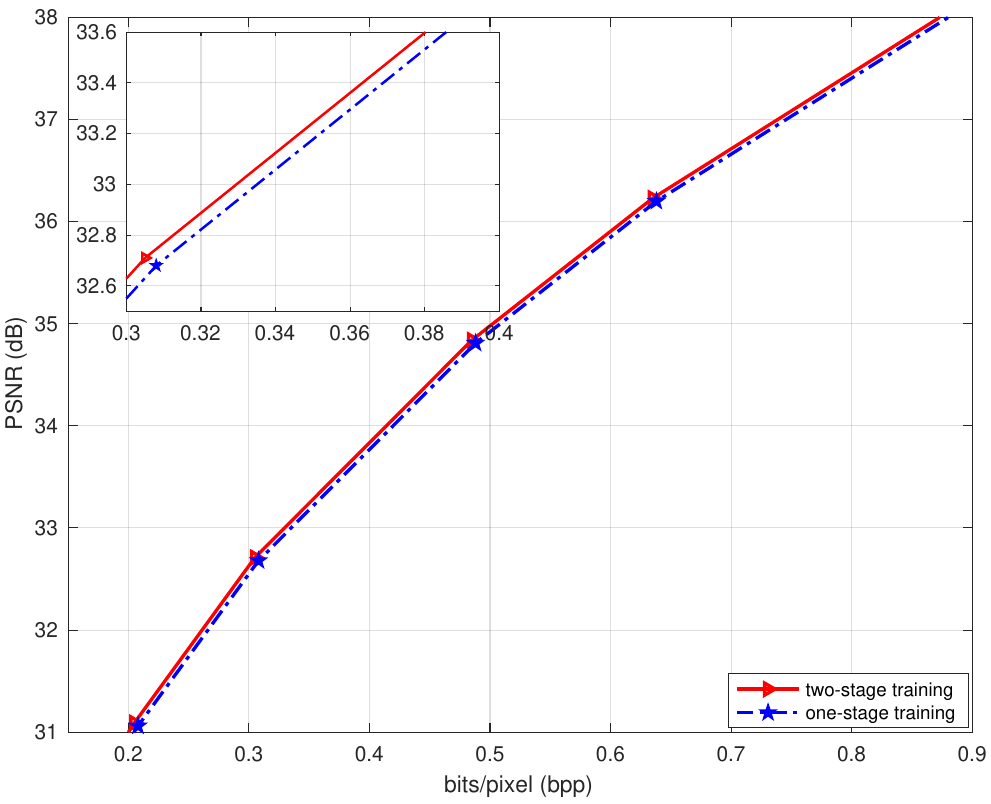}
\caption{Comparison between one-stage and two-stage trainings for the Kodak test set.}
\label{Figure_training_strategies}
\end{figure}

Fig. \ref{Figure_training_strategies} compares the R-D performance when the proposed two-stage training method and the simple one-stage training method is used on the Kodak test set respectively, which shows that the two-stage training method improves the performance by 0.02-0.03 dB without increasing complexity. Therefore it is beneficial to assign a larger weight to the LF components and a smaller weight to the LF components during the training.

\subsection{Applications of 3DM-WeConv Layers in Other Vision Tasks} 
\label{sec:weconv_applications}

To demonstrate the generality and effectiveness of the proposed 3DM-WeConv layer, we apply it to some other computer vision tasks, including video compression, image classification, image segmentation, and image denoising.

\subsubsection{Video Compression}

We first integrate the 3DM-WeConv layer into a learned deep video compression framework named DVC \cite{Lu_2019_CVPR}. We replace the downsampling layers in DVC by the 3DM-WeConv layers in the residual encoder, residual decoder, motion encoder, and motion decoder networks in \cite{Lu_2019_CVPR}. 

\begin{figure}[!t]
\centering
\includegraphics[scale=0.4]{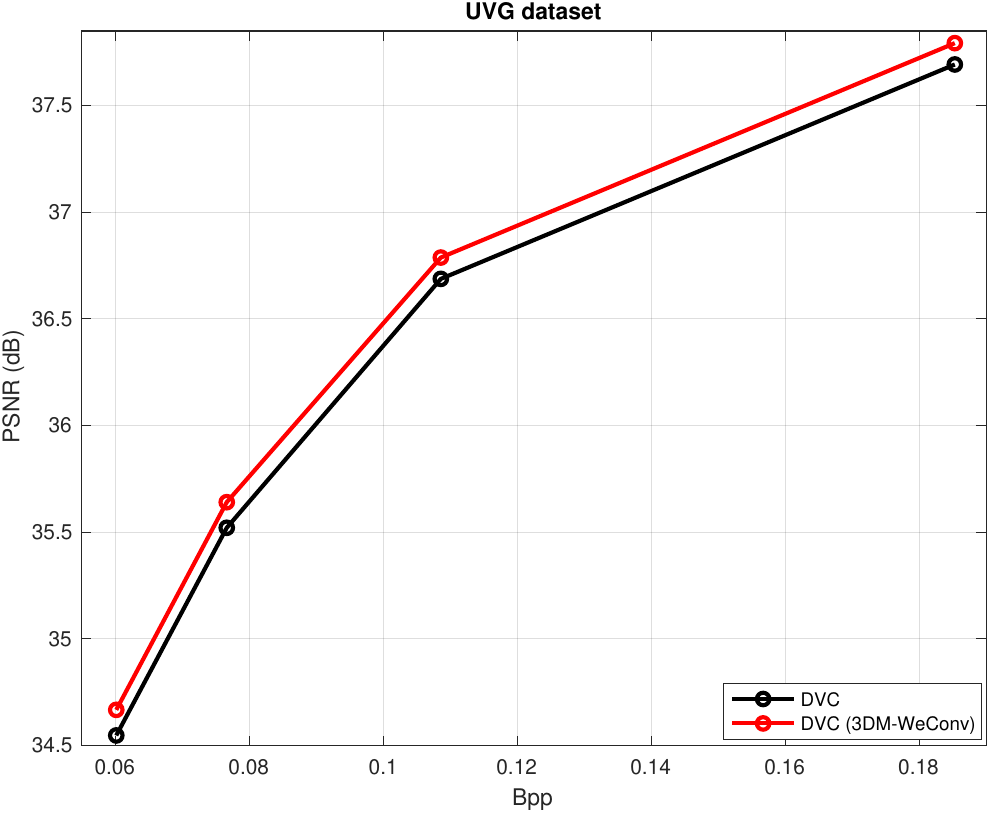}
\caption{Application of 3DM-WeConv in the deep video coding scheme DVC \cite{Lu_2019_CVPR} (UVG test set).}
\label{UVG_plot}
\end{figure}

Figure~\ref{UVG_plot} compares the R-D performance of the original DVC and the 3DM-WeConv-based DVC on the UVG test set \cite{UVG_dataset}, which shows that 3DM-WeConv can achieve an improvement of 0.15 dB at all bit rates. Further gains can be obtained if we also apply the 3DWeChARM in the entropy coding of the DVC scheme.

\subsubsection{Image Classification}

Next, we apply the 3DM-WeConv layer to image classification by replacing the standard downsampling modules in the ResNet-18 and ResNet-34~\cite{resblock} networks by the 3DM-WeConv layers. The models are then trained and evaluated on the CIFAR-100~\cite{cifar100} dataset. As reported in Table~\ref{tab:classification_results}, the 3DM-WeConv-based models achieve higher classification accuracy than the original models, demonstrating 3DM-WeConv's ability to capture discriminative features effectively.

\begin{table}[!t] 
\caption{Application of 3DM-WeConv Layer in image classification (CIFAR-100~\cite{cifar100} dataset).} 
\label{tab:classification_results} 
\centering 
\begin{tabular}{ccc} 
\hline 
\textbf{Model} & \textbf{Top 1 Acc. (\%)} & \textbf{Top 5 Acc. (\%)}\\ 
\hline
ResNet-18  ~\cite{resblock} & 75.61 & 93.05\\
ResNet-18 + 3DM-WeConv & \textbf{76.24} & \textbf{93.36}\\
\hline 
ResNet-34 ~\cite{resblock} & 76.96 & 0.934\\
ResNet-34 + 3DM-WeConv & \textbf{77.68} &\textbf{0.952}\\
\hline 
\end{tabular} 
\end{table}

\subsubsection{Image Segmentation}

In this example, we apply the 3DM-WeConv layer to image segmentation, by replacing the downsampling modules in the UperNet50 method ~\cite{zhou2018semantic} with the 3DM-WeConv layers. 

Table~\ref{tab:segmentation_results} demonstrates the performance on the ADE20K test set~\cite{ADE20K_dataset} , which shows 3DM-WeConv can improve the mean Intersection over Union (mIoU) and pixel accuracy.

\begin{table}[!t] 
\caption{Application of 3DM-WeConv in Image segmentation (ADE20K test set).} 
\label{tab:segmentation_results} 
\centering 
\begin{tabular}{ccc} 
\hline 
\textbf{Model} & \textbf{mIoU (\%)} & \textbf{Pixel Accuracy (\%)} \\
\hline 
UperNet50 ~\cite{zhou2018semantic} & 40.44 & 79.80 \\
UperNet50 + 3DM-WeConv & \textbf{41.24} & \textbf{80.26} \\
\hline 
\end{tabular} 
\end{table}

\subsubsection{Image Denoising}

In this example, we evaluate the performance of the 3DM-WeConv layer in image denoising. We replace the upsampling and downsampling modules in the MIRNet-v2 method ~\cite{zamir2023learning} by the 3DM-WeConv layers. The performance is measured using two widely used test sets: the SIDD test set~\cite{abdelhamed2018denoising} and the DND test set~\cite{plotz2017benchmarking}. 

As shown in Table~\ref{tab:denoising_results}, 3DM-WeConv also yields improved denoising performance in both PSNR and SSIM.

\begin{table}[!t]
\caption{Application of 3DM-WeConv in Image denoising (the SIDD~\cite{abdelhamed2018denoising} and DND~\cite{plotz2017benchmarking} test sets).}
\label{tab:denoising_results}
\centering
\begin{tabular}{cccc}
\hline
\textbf{Model} & \textbf{Test Set} & \textbf{PSNR (dB)} & \textbf{SSIM} \\
\hline
MIRNet-v2~\cite{zamir2023learning} & SIDD & 39.84 & 0.959 \\
MIRNet-v2 + 3DM-WeConv & SIDD & \textbf{40.09} & \textbf{0.961} \\
\hline
MIRNet-v2~\cite{zamir2023learning} & DND  & 39.86 & 0.955 \\
MIRNet-v2 + 3DM-WeConv & DND  & \textbf{40.07} & \textbf{0.958} \\
\hline
\end{tabular}
\end{table}

\section{Conclusions}

In this paper, We develop an improved and effective approach to integrate the classic 3D and multi-level discrete wavelet transforms (DWT) into both the convolutional layers and the entropy coding stage of learned image compression (LIC), based on our previous work in \cite{Fu_EECV}. By transforming latent representations into the 3D DWT domain, we can enhance the sparsity of the data in the frequency domain across different channels and within each channel. This can effectively improve the R-D performance with negligible complexity. 

On the Kodak, Tecnick 100, and CLIC test sets, our new framework achieves BD-Rate reductions of $-12.24\%$, $-15.51\%$, and $-12.97\%$ relative to H.266/VVC, and reductions of $-2.34\%$, $-2.34\%$, and $-3.54\%$ compared to our previous work in \cite{Fu_EECV}. Our model also reduces the number of model parameters by more than $30\%$. Therefore, our method achieves an excellent trade-off among performance, model complexity, and decoding time compared to other leading LIC methods.

This work bridges the important gap between the traditional DSP-based image compression and the latest learning-based approach, and will enable more wavelet theories to be applied in learning-based compression.  The proposed 3DWeChARM entropy coding can also be used in transformer-based image/video compression schemes. 

More importantly, our ablation studies confirm that the proposed 3DM-WeConv layer can also be applied in many other computer vision tasks. These findings highlight the potential of DWT as a powerful, universal, and low-complexity tool for other deep learning applications.

\bibliographystyle{IEEEtran}
%\bibliography{egbib}{}

\begin{IEEEbiography}
[{\includegraphics[width=1in,height=1.25in,clip,keepaspectratio]{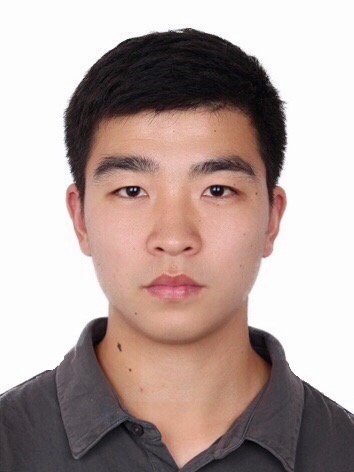}}]{Haisheng Fu} (Student Member, IEEE) received the Ph.D. degree in electronic science and technology from Xi’an Jiaotong University, Xi’an. During his Ph.D. studies, he spent two years as a joint Ph.D. student at Simon Fraser University, Canada. He is currently a postdoctoral fellow at Simon Fraser University. His research interests include Machine Learning, Image and Video Compression, Deep Learning, and VLSI Design. He is also an active reviewer for several prestigious journals and conferences, including ICCV, IEEE TPAMI, IEEE TIP, IEEE TCSVT, ICASSP, and ICIP.
\end{IEEEbiography}

\begin{IEEEbiography}[{\includegraphics[width=1in,height=1.25in,clip,keepaspectratio]{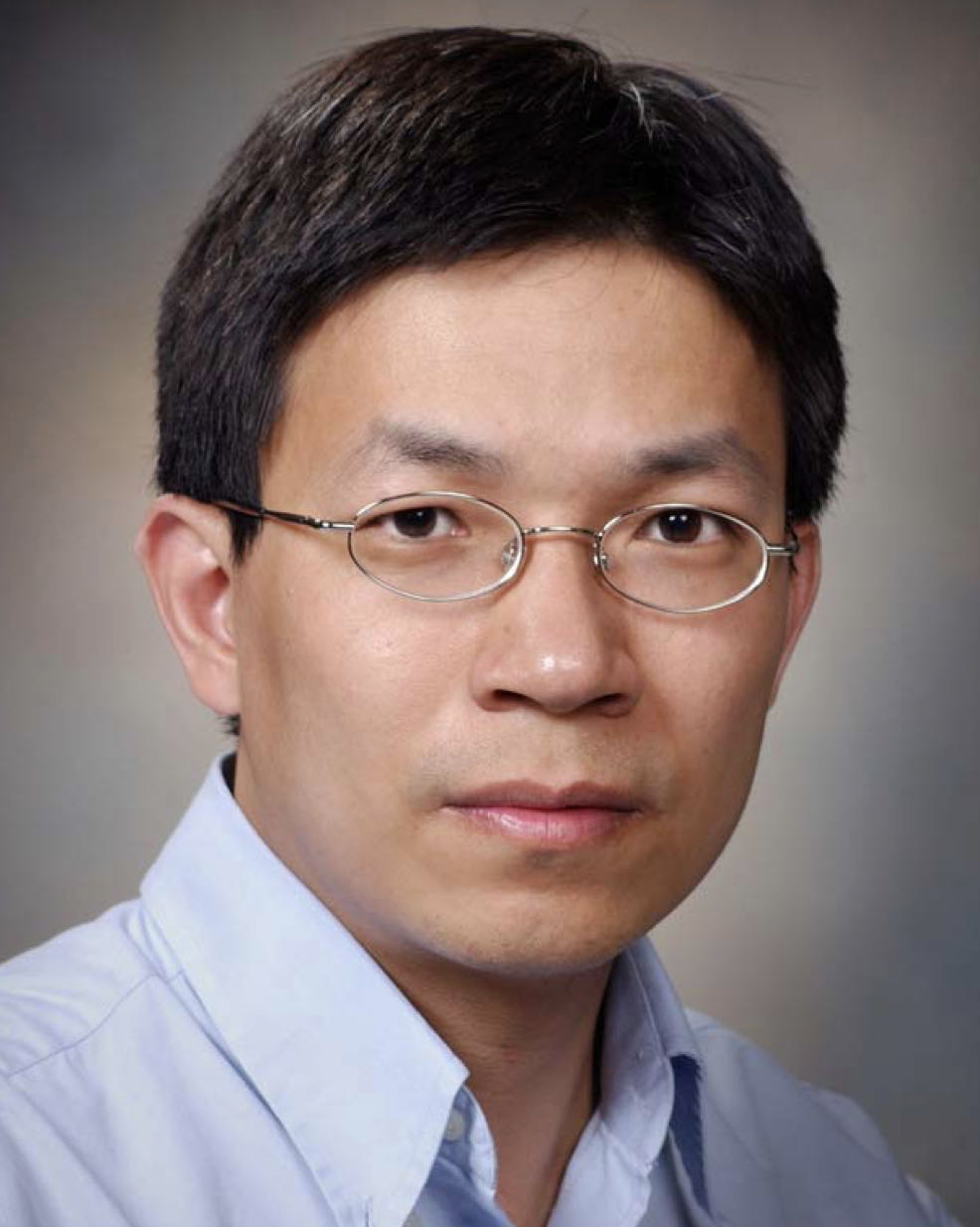}}]{Jie Liang} (Senior Member, IEEE)  received the B.E. and M.E. degrees from Xi'an Jiaotong University, China, the M.E. degree from National University of Singapore, and the PhD degree from the Johns Hopkins University, USA, in 1992, 1995, 1998, and 2003, respectively. From 2003 to 2004, he worked at the Video Codec Group of Microsoft Digital Media Division. Since May 2004, he has been with the School of Engineering Science, Simon Fraser University, Canada, where he is currently a Professor.
 
Jie Liang's research interests include Image and Video Processing, Computer Vision, and Deep Learning. He had served as an Associate Editor for several journals, including IEEE Transactions on Image Processing, IEEE Transactions on Circuits and Systems for Video Technology (TCSVT), and IEEE Signal Processing Letters. He has also served on three IEEE Technical Committees. He received the 2014 IEEE TCSVT Best Associate Editor Award, 2014 SFU Dean of Graduate Studies Award for Excellence in Leadership, and 2015 Canada NSERC Discovery Accelerator Supplements (DAS) Award.
\end{IEEEbiography}

\begin{IEEEbiography}[{\includegraphics[width=1in,height=1.25in,clip,keepaspectratio]{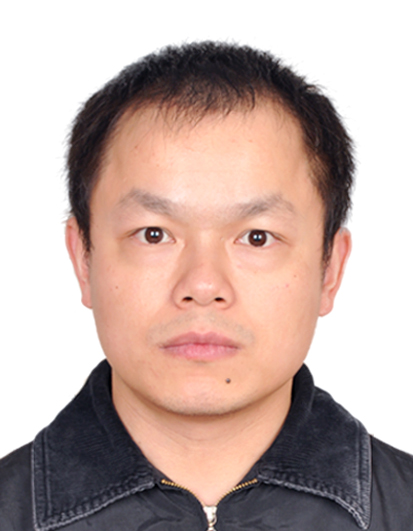}}]{Feng Liang} is currently Professor of the Microelectronics School at Xi'an Jiaotong University. He earned his B.E. from Zhengzhou University and his M.E. and Ph.D. from Xi'an Jiaotong University. His current research interests include Signal Processing, Machine Learning, VLSI design, CIM, and computer architecture.
\end{IEEEbiography}

\begin{IEEEbiography}
[{\includegraphics[width=1in,height=1.25in,clip,keepaspectratio]{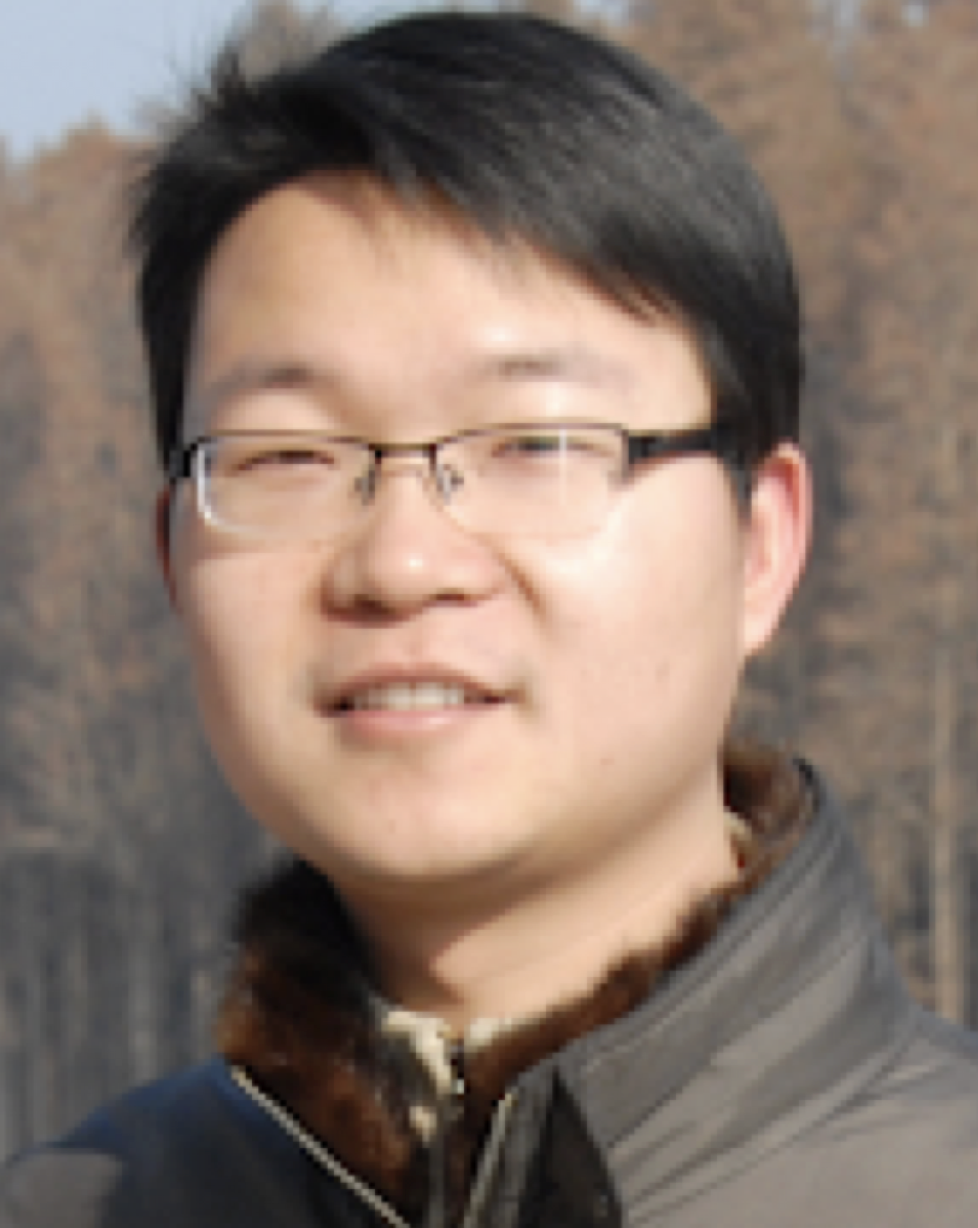}}]{Zhenman Fang} (Member, IEEE) received the Ph.D. degree in computer science from Fudan University, Shanghai, China, in 2014. He conducted postdoctoral research at the University of California, Los Angeles (UCLA), Los Angeles, CA, USA, from 2014 to 2017, and worked as a Staff Software Engineer at Xilinx, San Jose, CA, USA, from 2017 to 2019. He is currently an Assistant Professor with the School of Engineering Science, Simon Fraser University, Burnaby, BC, Canada. His recent research focuses on customizable computing with specialized hardware acceleration, including emerging application characterization and acceleration, novel accelerator-rich and near-data computing architecture designs, and corresponding programming, runtime, and tool support.
\end{IEEEbiography}

\begin{IEEEbiography}[{\includegraphics[width=1in,height=1.25in,clip,keepaspectratio]{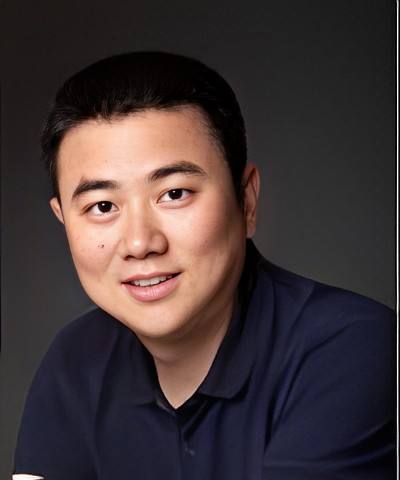}}]{Guohe Zhang} received the B.S. and Ph.D. degrees in electronics science and technology from Xi’an Jiaotong University, Shaanxi, China, in 2003 and 2008, respec- tively. He is currently an Associate Professor with the School of Microelectronics, Xi’an Jiaotong University. In 2009, he joined the School of Elec- tronic and Information Engineering, as a Lec- turer. He was promoted to an Associated Professor, in 2013. From 2009 to 2011, he had a three year’s
Postdoctoral Researcher with the School of Nuclear Science and Technology, Xi’an Jiaotong University. From February to May of 2013, he had a short term visiting to the University of Liverpool, U.K. His research interests fall in the area of semiconductor device physics and modeling, VLSI design and testing.
\end{IEEEbiography}

\begin{IEEEbiography}[{\includegraphics[width=1in,height=1.25in,clip,keepaspectratio]{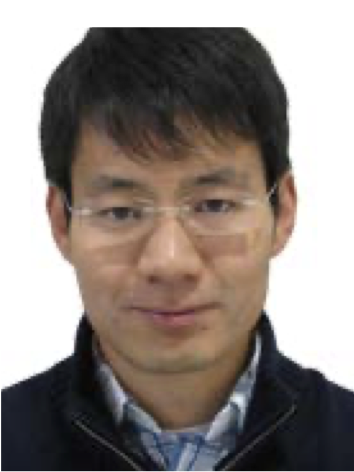}}]{Jingning Han} (Senior Member, IEEE) received the B.S. degree in electrical engineering from Tsinghua University, Beijing, China, in 2007, and the M.S. and Ph.D. degrees in electrical and computer engineering from the University of California at Santa Barbara, Santa Barbara, CA, USA, in 2008 and 2012, respectively.
He joined the WebM Codec Team, Google, Mountain View, CA, USA, in 2012, where he is the Main Architect of the VP9 and AV1 codecs, and leads the Software Video Codec Team. He has published more than 60 research articles. He holds more than 50 U.S. patents in the field of video coding. His research interests include video coding and computer science architecture. Dr. Han received the Dissertation Fellowship from the Department of Elec- trical and Engineering, University of California at Santa Barbara, in 2012. He was a recipient of the Best Student Paper Award at the IEEE International Conference on Multimedia and Expo, in 2012. He also received the IEEE Signal Processing Society Best Young Author Paper Award, in 2015.
\end{IEEEbiography}

\end{document}